\documentclass[10pt,twocolumn,letterpaper]{article}
 
\usepackage{iccv}
\usepackage{times}
\usepackage{epsfig}
\usepackage{graphicx}
\usepackage{amsmath}
\usepackage{amssymb}
\usepackage{subfig}
\usepackage{spverbatim}
\usepackage{etoolbox}
\makeatletter
\preto{\@verbatim}{\topsep=1pt \partopsep=1pt}
\makeatother

\usepackage{booktabs}
\usepackage[x11names,dvipsnames,table]{xcolor} %for use in color links
\usepackage{multirow}
\usepackage{array}
\newcolumntype{L}[1]{>{\raggedright\let\newline\\\arraybackslash\hspace{0pt}}m{#1}}
\newcolumntype{C}[1]{>{\centering\let\newline\\\arraybackslash\hspace{0pt}}m{#1}}
\newcolumntype{R}[1]{>{\raggedleft\let\newline\\\arraybackslash\hspace{0pt}}m{#1}}

\newcommand{\scatterwidth}{8.0cm}

\newcommand{\minipagewidthA}{8.0cm}
\newcommand{\minipagewidthB}{8.0cm}
\newcommand{\colwidth}{1.0cm}

\usepackage[linesnumbered,ruled]{algorithm2e}

\usepackage[breaklinks=true,bookmarks=false]{hyperref}

\iccvfinalcopy % *** Uncomment this line for the final submission

\def\iccvPaperID{****} % *** Enter the ICCV Paper ID here

\begin{document}

%%%%%%%%% TITLE
\title{Adversarial Examples for Semantic Segmentation and Object Detection}

\author{Cihang Xie\textsuperscript{1*}, Jianyu Wang\textsuperscript{2*}, Zhishuai Zhang\textsuperscript{1}\thanks{The first three authors contributed equally to this work. This work was done when Jianyu Wang was a Ph.D. student at UCLA}  ,
Yuyin Zhou\textsuperscript{1}, Lingxi Xie\textsuperscript{1}, Alan Yuille\textsuperscript{1} \\
\textsuperscript{1}Department of Computer Science, The Johns Hopkins University, Baltimore, MD 21218 USA \\
\textsuperscript{2}Baidu Research USA, Sunnyvale, CA 94089 USA \\
%\textsuperscript{3}Center for Imaging Science, The Johns Hopkins University, Baltimore, MD, USA \\
\{cihangxie306, wjyouch,  zhshuai.zhang, zhouyuyiner, 198808xc, alan.l.yuille\}@gmail.com}

\maketitle
%\thispagestyle{empty}

%%%%%%%%% ABSTRACT
\begin{abstract}
It has been well demonstrated that adversarial examples, i.e., natural images with visually imperceptible perturbations added,
cause deep networks to fail on image classification.
In this paper, we extend adversarial examples to semantic segmentation and object detection which are much more difficult.
Our observation is that both segmentation and detection are based on classifying multiple targets on an image
(e.g., the target is a pixel or a receptive field in segmentation, and an object proposal in detection).
This inspires us to optimize a loss function over a set of pixels/proposals for generating adversarial perturbations.
Based on this idea, we propose a novel algorithm named Dense Adversary Generation (DAG),
which generates a large family of adversarial examples,
and applies to a wide range of state-of-the-art deep networks for segmentation and detection.
We also find that the adversarial perturbations can be transferred across networks with different training data,
based on different architectures, and even for different recognition tasks.
In particular, the transferability across networks with the same architecture is more significant than in other cases.
Besides, summing up heterogeneous perturbations often leads to better transfer performance,
which provides an effective method of black-box adversarial attack.
\end{abstract}

%%%%%%%%% BODY TEXT
\section{Introduction}
\label{Introduction}

\begin{figure}[t!]
\centering
\subfloat{\includegraphics[width=0.48\columnwidth]{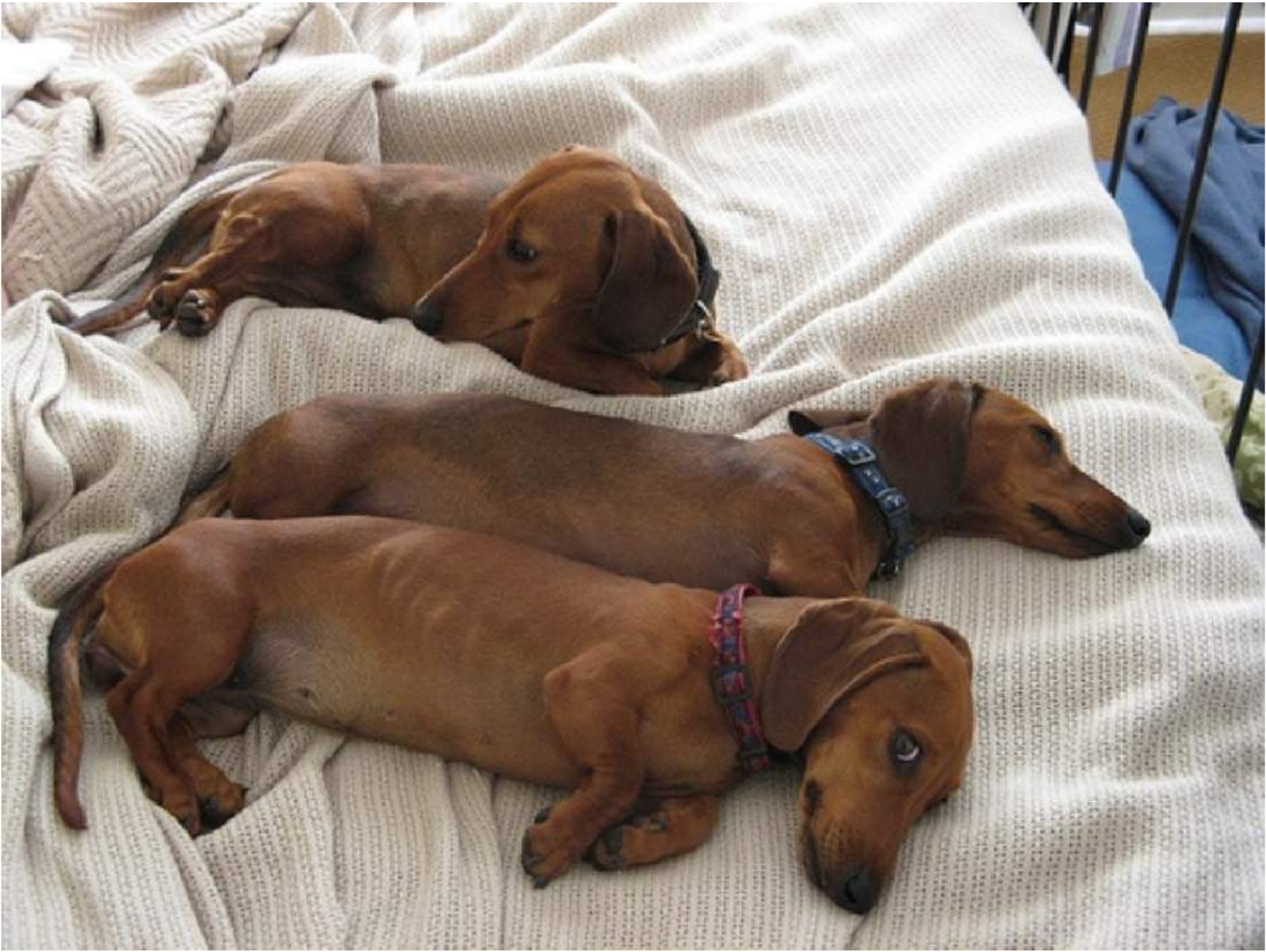}}
\hfill
\subfloat{\includegraphics[width=0.48\columnwidth]{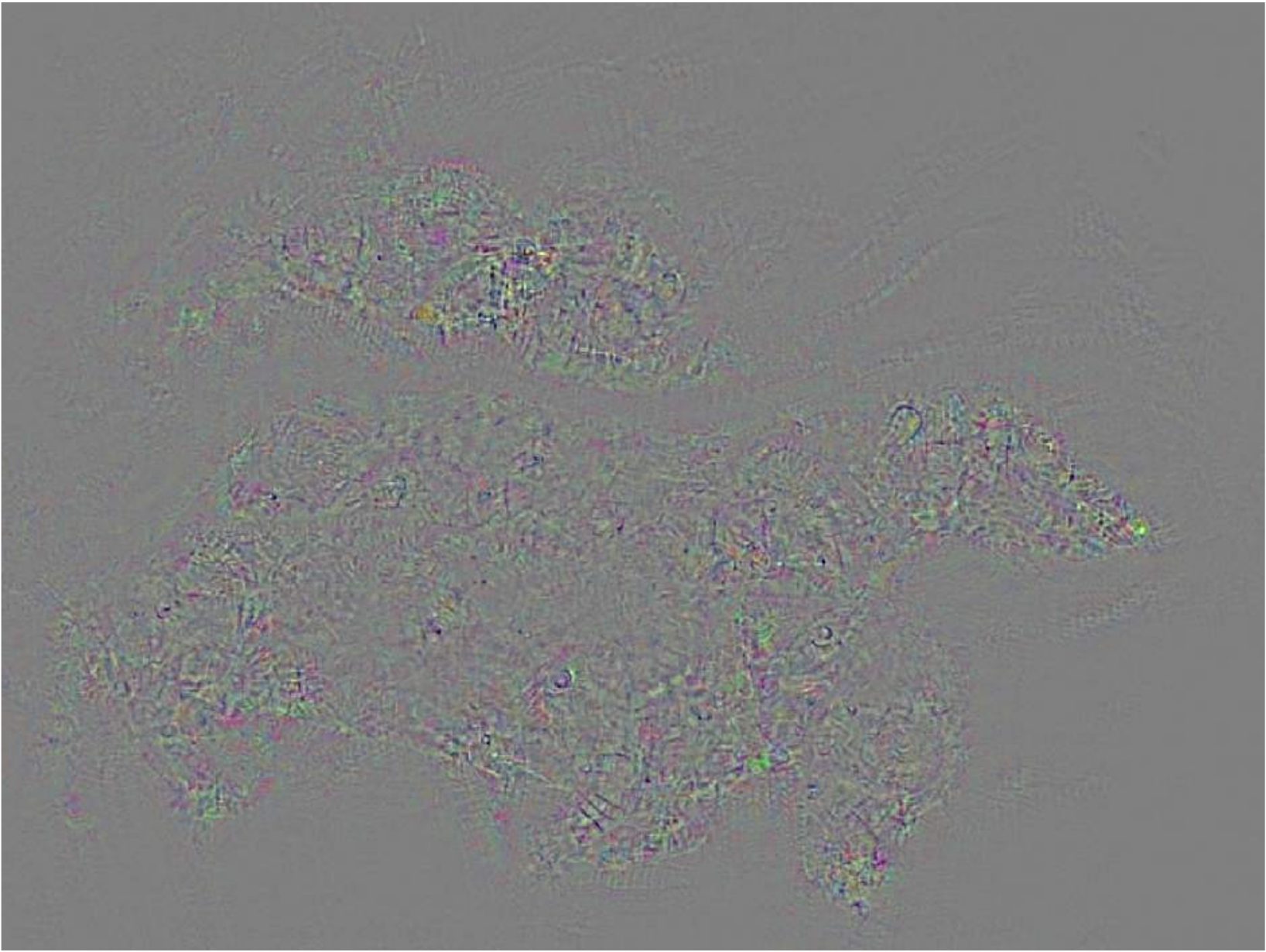}}
\hfill
\subfloat{\includegraphics[width=0.48\columnwidth]{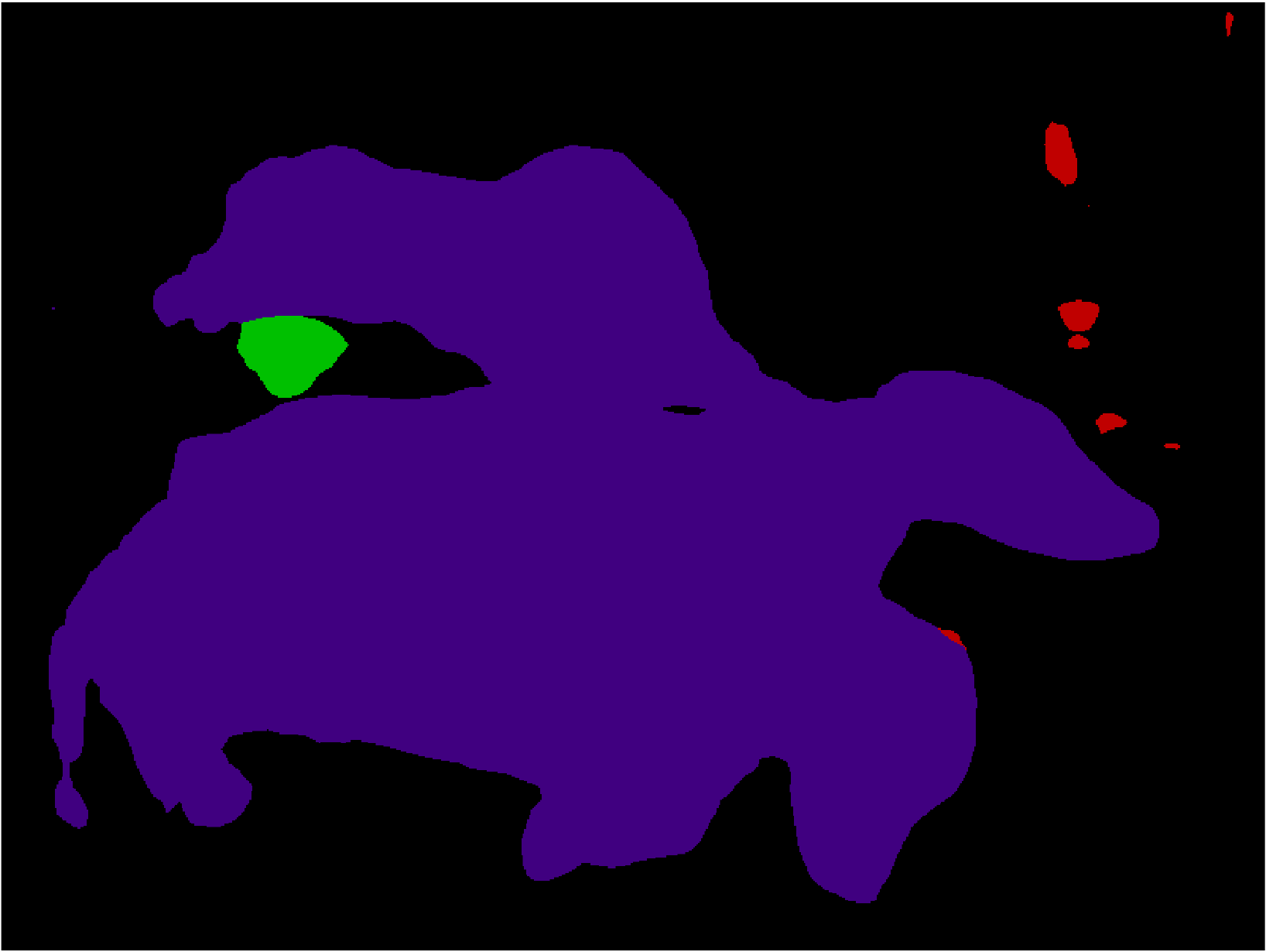}}
\hfill
\subfloat{\includegraphics[width=0.48\columnwidth]{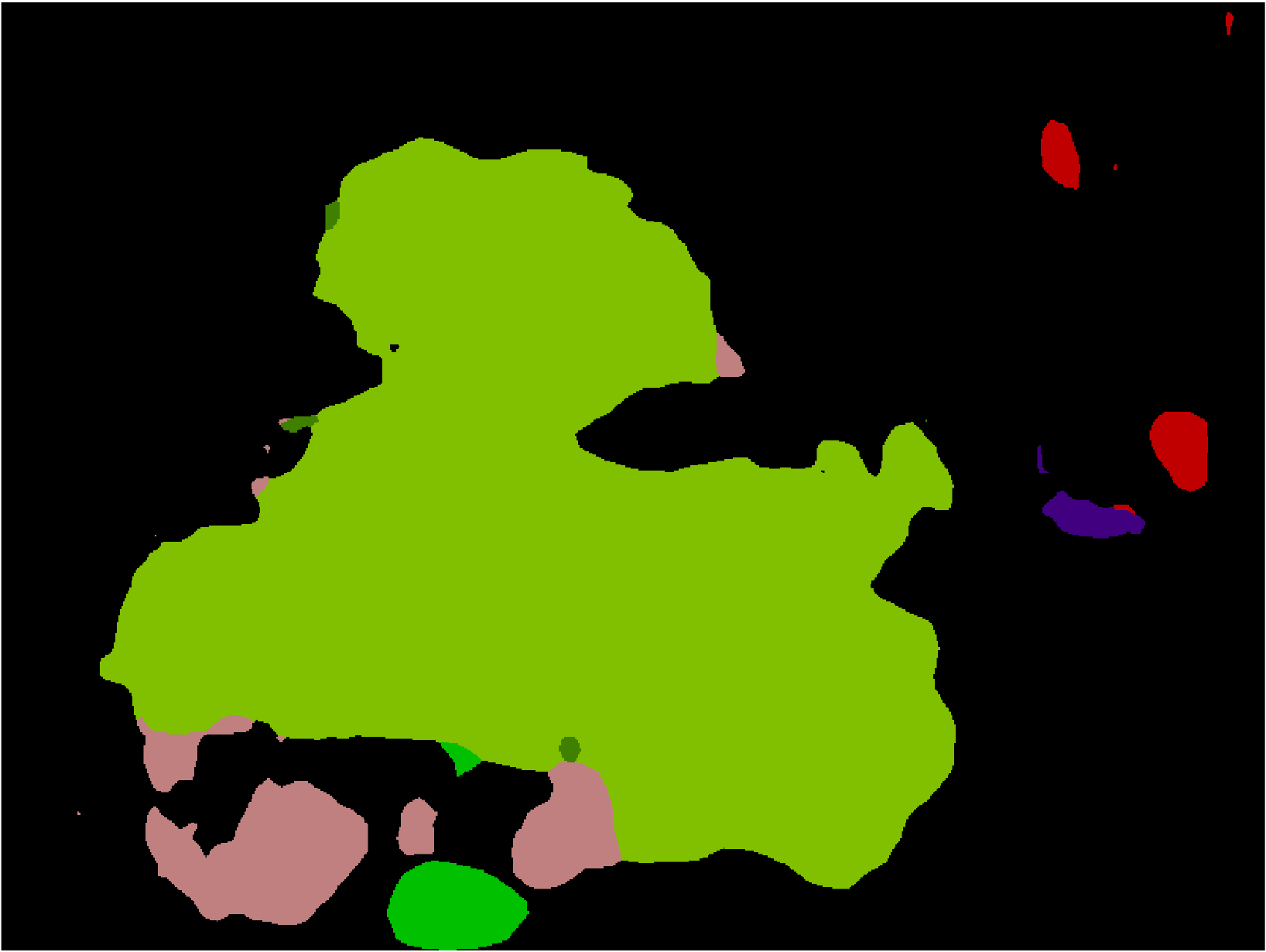}}
\hfill
\subfloat{\includegraphics[width=0.48\columnwidth]{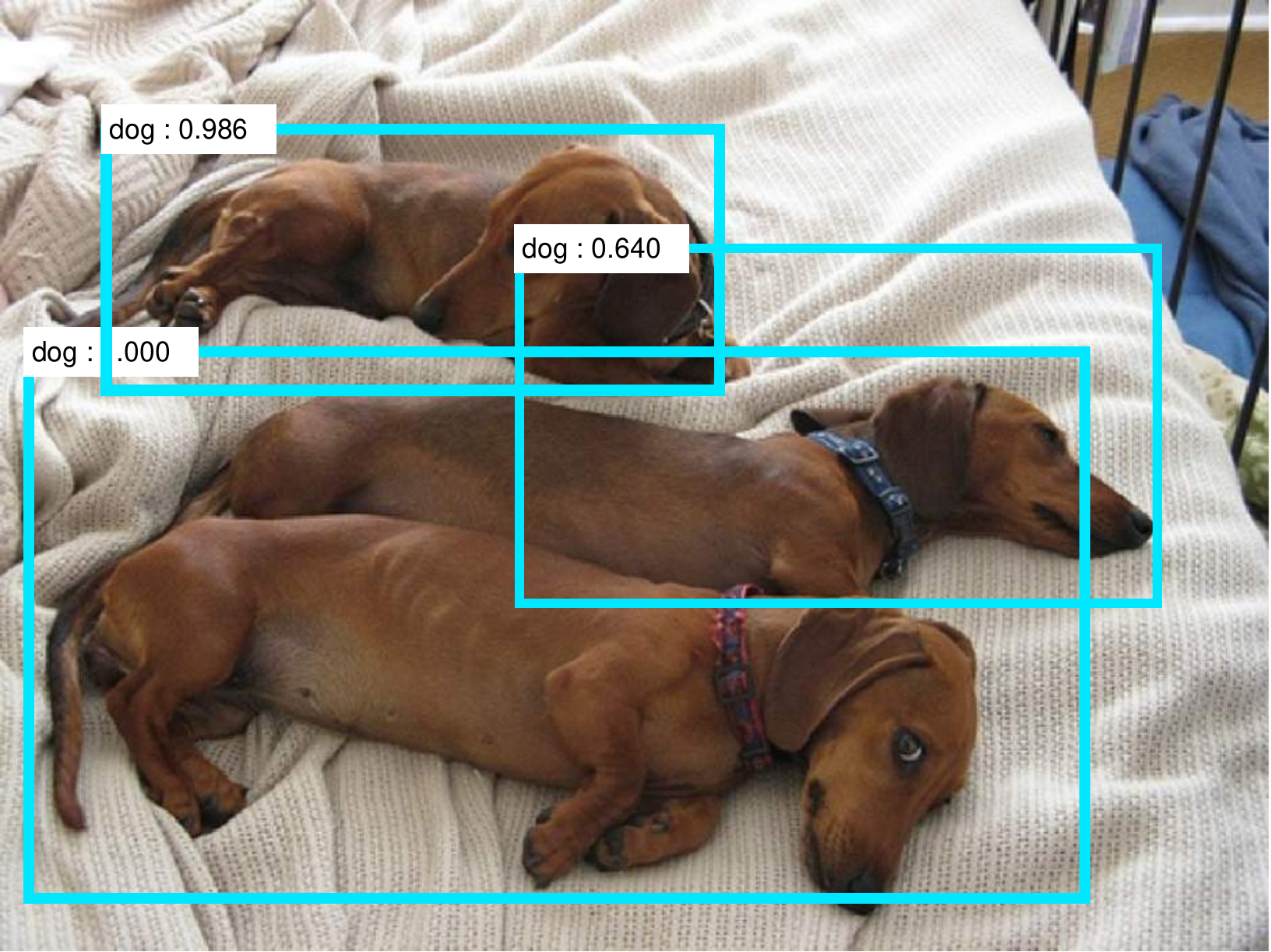}}
\hfill
\subfloat{\includegraphics[width=0.48\columnwidth]{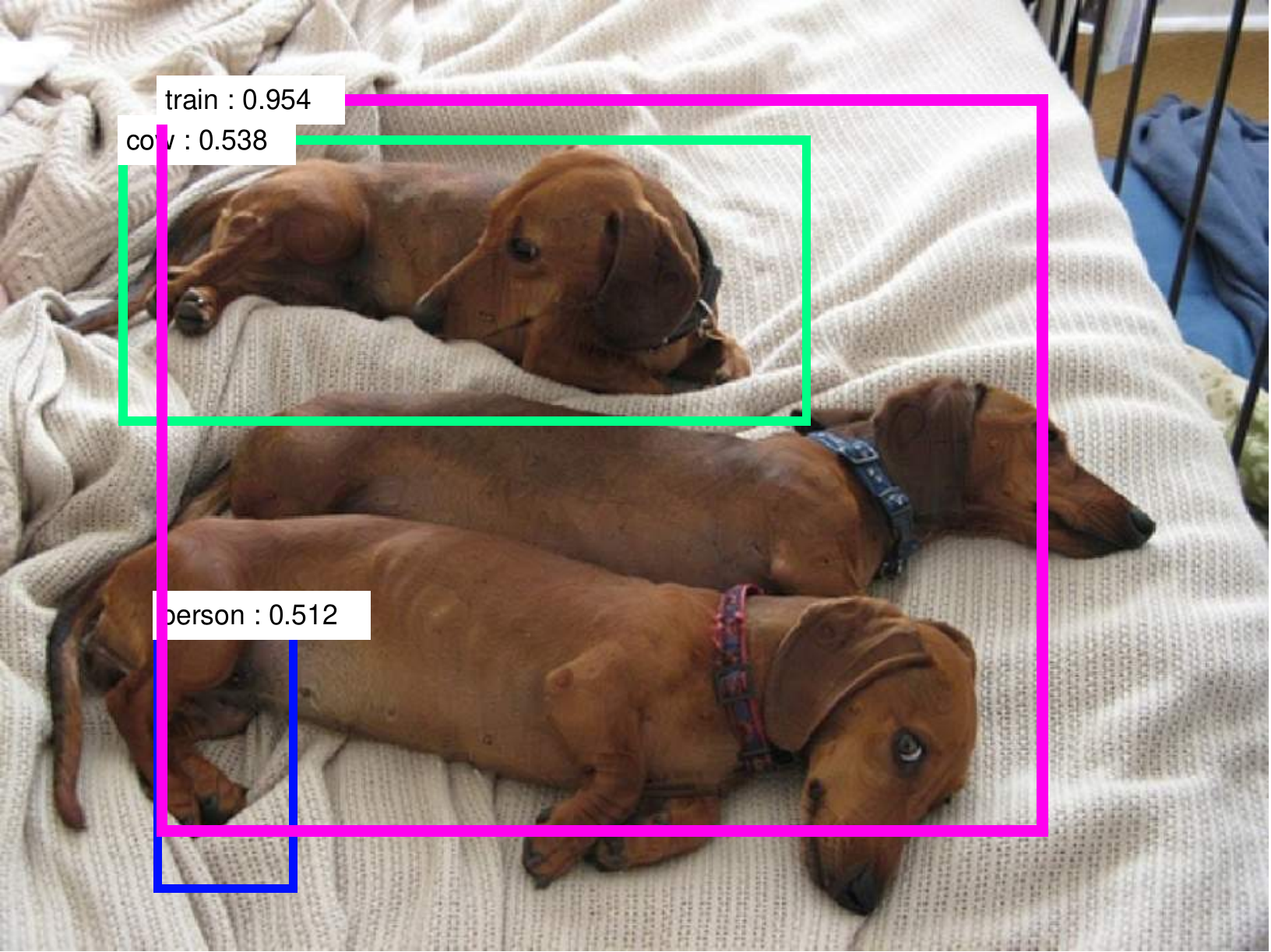}}
\hfill
\caption{
    An Adversarial example for semantic segmentation and object detection.
    FCN~\cite{long2015fully} is used for segmentation, and Faster-RCNN~\cite{ren2015faster} is used for detection.
    Left column: the original image (top row)
    with the normal segmentation (the purple region is predicted as {\em dog}) and detection results.
    Right column: after the adversarial perturbation (top row, {\bf magnified by $\mathbf{10}$}) is added to the original image,
    both segmentation (the light green region as {\em train} and the pink region as {\em person})
    and detection results are completely wrong.
    Note that, though the added perturbation can confuse both networks,
    it is visually imperceptible (the maximal absolute intensity in each channel is less than $10$).
}
\vspace{-0.5cm}
\label{Fig:AdversarialExamples}
\end{figure}

Convolutional Neural Networks
(CNN)~\cite{krizhevsky2012imagenet}\cite{simonyan2015very}\cite{szegedy2015going}\cite{he2016deep}
have become the state-of-the-art solution for a wide range of visual recognition problems.
Based on a large-scale labeled dataset such as {\bf ImageNet}~\cite{deng2009imagenet}
and powerful computational resources like modern GPUs,
it is possible to train a hierarchical deep network to capture different levels of visual patterns.
A deep network is also capable of generating transferrable features
for different tasks such as image classification~\cite{donahue2014decaf} and instance retrieval~\cite{sharif2014cnn},
or being fine-tuned to deal with a wide range of vision tasks,
including object detection~\cite{girshick2014rich}\cite{girshick2015fast},  visual concept discovery~\cite{wang2015discovering},
semantic segmentation~\cite{long2015fully}\cite{chen2016deeplab}\cite{zheng2015conditional}, boundary detection~\cite{xie2015holistically}\cite{shen2015deepcontour}, etc.

Despite their success in visual recognition and feature representation,
deep networks are often sensitive to small perturbations to the input image.
In~\cite{szegedy2013intriguing},
it was shown that adding visually imperceptible perturbations can result in failures for image classification.
These perturbed images, often called {\em adversarial examples},
are considered to fall on some areas in the large, high-dimensional feature space which are not explored in the training process.
Thus, investigating this not only helps understand the working mechanism of deep networks,
but also provides opportunities to improve the robustness of network training.

In this paper, we go one step further by generating adversarial examples for semantic segmentation and object detection, and showing the transferability of them. To the best of our knowledge, this topic has not been systematically studied ({\em e.g.}, on a large dataset) before. Note that these tasks are much more difficult,
as we need to consider orders of magnitude more targets ({\em e.g.}, pixels or proposals).
Motivated by the fact that each target undergoes a separate classification process,
we propose the Dense Adversary Generation ({\bf DAG}) algorithm,
which considers all the targets simultaneously and optimizes the overall loss function.
The implementation of DAG is simple,
as it only involves specifying an adversarial label for each target and performing iterative gradient back-propagation.
In practice, the algorithm often comes to an end after a reasonable number of, say, $150$ to $200$, iterations.
Figure~\ref{Fig:AdversarialExamples} shows an  adversarial example which can confuse both deep segmentation and detection networks.

We point out that generating an adversarial example is more difficult in detection than in segmentation,
as the number of targets is orders of magnitude larger in the former case, {\em e.g.}, for an image with $K$ pixels,
the number of possible proposals is $O\!\left(K^2\right)$ while the number of pixels is only $O\!\left(K\right)$,
where $O\!\left(\cdot\right)$ is the big-O notation.
In addition, if only a subset of proposals are considered,
the perturbed image may still be correctly recognized after a new set of proposals are extracted
(note that DAG aims at generating recognition failures on the original proposals).
To increase the robustness of adversarial attack,
we change the intersection-over-union (IOU) rate to preserve an increased but still reasonable number of proposals in optimization.
In experiments, we verify that when the proposals are dense enough on the original image,
it is highly likely that incorrect recognition results are also produced on the new proposals generated on the perturbed image.
We also study the effectiveness and efficiency of the algorithm
with respect to the {\em denseness} of the considered proposals.

Following~\cite{szegedy2013intriguing}, we investigate the transferability of the generated perturbations.
To this end, we use the adversarial perturbation computed on one network to attack another network.
Three situations are considered:
(1) networks with the same architecture but trained with different data;
(2) networks with different architectures but trained for the same task; and (3) networks for different tasks.
Although the difficulty increases as the difference goes more significant,
the perturbations generated by DAG is able to transfer to some extent.
Interestingly, adding two or more heterogeneous perturbations significantly increases the transferability,
which provides an effective way of performing black-box adversarial attack~\cite{papernot2016practical}
to some networks with unknown structures and/or properties.

The remainder of this paper is organized as follows.
Section~\ref{RelatedWork} briefly introduces prior work related to our research.
Section~\ref{Generating} describes our algorithm for generating adversarial perturbations,
and Section~\ref{Transferring} investigates the transferability of the perturbations.
Conclusions are drawn in Section~\ref{Conclusions}.

\section{Related Work}
\label{RelatedWork}

\subsection{Deep Learning for Detection and Segmentation}
\label{RelatedWork:DeepLearning}

Deep learning approaches, especially deep convolutional neural networks,
have been very successful in object detection~\cite{ren2015faster}\cite{li2016r}\cite{lin2016feature}
and semantic segmentation~\cite{long2015fully}\cite{chen2016deeplab} tasks.
Currently, one of the most popular object detection pipeline~\cite{ren2015faster}\cite{li2016r}\cite{lin2016feature}
involves first generating a number of proposals of different scales and positions,
classifying each of them, and performing post-processing such as non-maximal suppression (NMS).
On the other hand, the dominating segmentation pipeline~\cite{long2015fully}
works by first predicting a class-dependent score map at a reduced resolution,
and performing up-sampling to obtain high-resolution segmentation.
~\cite{chen2016deeplab} incorporates the ``atrous'' algorithm and the conditional random field (CRF) to this pipeline
to improve the segmentation performance further.

\subsection{Adversarial Attack and Defense}
\label{RelatedWork:Adversarial}

Generating adversarial examples for classification has been extensively studied in many different ways recently.
\cite{szegedy2013intriguing} first showed that adversarial examples,
computed by adding visually imperceptible perturbations to the original images,
make CNNs predict a wrong label with high confidence.
\cite{goodfellow2014explaining} proposed a simple and fast gradient sign method to generate adversarial examples based on the linear nature of CNNs.
\cite{moosavi2015deepfool} proposed a simple algorithm to compute the minimal adversarial perturbation
by assuming that the loss function can be linearized around the current data point at each iteration.  \cite{moosavi2016universal} showed the existence of universal (image-agnostic) adversarial perturbations. \cite{baluja2017adversarial} trained a network to generate adversarial examples for a particular target model (without using gradients).
\cite{kurakin2016adversarial} showed the adversarial examples for machine learning systems also exist in the physical world.
\cite{liu2016delving} studied the transferability of both non-targeted and targeted adversarial examples, and proposed an ensemble-based approaches to generate adversarial examples with stronger transferability. \cite{nguyen2015deep} generated images using evolutionary algorithms that are unrecognizable to humans,
but cause CNNs to output very confident (incorrect) predictions.
This can be thought of as in the opposite direction of above works.

In contrast to generating adversarial examples,
there are some works trying to reduce the effect of adversarial examples. \cite{lou2016foveation} proposed a forveation-based mechanism to alleviate adversarial examples. \cite{papernot2016distillation} showed networks trained using defensive distillation can effectively against adversarial examples, while \cite{carlini2016towards} developed stronger attacks which are unable to defend by defensive distillation.
\cite{kurakin2016scale} trained the network on adversarial examples using the large-scale ImageNet, and showed that this brings robustness to adversarial attack. This is imporved by \cite{tramer2017ensemble}, which proposed an ensemble adversarial training method to increase the network robustness to black-box attacks. \cite{metzen2017detecting} trained a detector on the inner layer of the classifier to detect adversarial examples.

There are two concurrent works \cite{fischer2017adversarial} and \cite{metzen2017universal} that studied adversarial examples in semantic segmentation on the Cityscapes dataset~\cite{cordts2016cityscapes}, where \cite{fischer2017adversarial} showed the existence of adversarial examples, and \cite{metzen2017universal} showed the existence of universal perturbations. We refer interested readers to their papers for details.

\section{Generating Adversarial Examples}
\label{Generating}

In this section, we introduce DAG algorithm.
Given an image and the recognition targets (proposals and/or pixels),
DAG generates an adversarial perturbation which is aimed at confusing as many targets as possible.

\subsection{Dense Adversary Generation}
\label{Generating:Algorithm}

Let $\mathbf{X}$ be an image which contains $N$ recognition {\em targets} ${\mathcal{T}}={\left\{t_1,t_2,\ldots,t_N\right\}}$.
Each target $t_n$, ${n}={1,2,\ldots,N}$, is assigned a ground-truth class label ${l_n}\in{\left\{1,2,\ldots,C\right\}}$,
where $C$ is the number of classes,
{\em e.g.}, ${C}={21}$ (including the {\em background} class) in the {\bf PascalVOC} dataset~\cite{everingham2007pascal}.
Denote ${\mathcal{L}}={\left\{l_1,l_2,\ldots,l_n\right\}}$.
The detailed form of $\mathcal{T}$ varies among different tasks.
In image classification, $\mathcal{T}$ only contains one element, {\em i.e.}, the entire image.
Conversely, $\mathcal{T}$ is composed of all pixels (or the corresponding receptive fields) in semantic segmentation,
and all proposals in object detection.
We will discuss how to construct $\mathcal{T}$ in Section~\ref{Generating:Selection}.

Given a deep network for a specific task,
we use ${\mathbf{f}\!\left(\mathbf{X},t_n\right)}\in{\mathbb{R}^C}$
to denote the classification score vector (before softmax normalization) on the $n$-th recognition target of $\mathbf{X}$.
To generate an adversarial example, the goal is to make the predictions of all targets go wrong,
{\em i.e.}, $\forall n$, ${\arg\max_c\left\{f_c\!\left(\mathbf{X}+\mathbf{r},t_n\right)\right\}}\neq{l_n}$.
Here $\mathbf{r}$ denotes an adversarial perturbation added to $\mathbf{X}$.
To this end, we specify an adversarial label $l'_n$ for each target,
in which $l'_n$ is randomly sampled from other incorrect classes,
{\em i.e.}, ${l'_n}\in{\left\{1,2,\ldots,C\right\}\setminus\left\{l_n\right\}}$.
Denote ${\mathcal{L}'}={\left\{l'_1,l'_2,\ldots,l'_n\right\}}$.
In practice, we define a random permutation function ${\pi}:{\left\{1,2,\ldots,C\right\}\rightarrow\left\{1,2,\ldots,C\right\}}$ for every image independently,
in which ${\pi\!\left(c\right)}\neq{c}$ for ${c}={1,2,\ldots,C}$,
and generate $\mathcal{L}'$ by setting ${l'_n}={\pi\!\left(l_n\right)}$ for all $n$.
Under this setting, the loss function covering all targets can be written as:
\begin{equation}
\label{Eqn:LossFunction}
{L\!\left(\mathbf{X},\mathcal{T},\mathcal{L},\mathcal{L}'\right)}=
    {{\sum_{n=1}^N}\left[{f}_{l_n}\!\left(\mathbf{X},t_n\right)-{f}_{l_n'}\!\left(\mathbf{X},t_n\right)\right]}
\end{equation}
Minimizing $L$ can be achieved via making every target to be incorrectly predicted,
{\em i.e.}, suppressing the confidence of the original correct class $f_{l_n}\!\left(\mathbf{X}+\mathrm{r},t_n\right)$,
while increasing that of the desired (adversarial) incorrect class $f_{l'_n}\!\left(\mathbf{X}+\mathrm{r},t_n\right)$.

\setlength{\algomargin}{0.7em}
%\SetAlgoNlRelativeSize{0.1}
\SetNlSkip{0.4em}
\SetNlSty{text}{}{}
\begin{algorithm}[t!]
\SetKwInOut{Input}{Input}
\SetKwInOut{Output}{Output}
\SetKwInOut{Return}{Return}
\Input{
    input image $\mathbf{X}$;\\
    \ the classifier ${\mathbf{f}\!\left(\cdot,\cdot\right)}\in{\mathbb{R}^C}$;\\
    \ the target set ${\mathcal{T}}={\left\{t_1,t_2,\ldots,t_N\right\}}$;\\
    \ the original label set ${\mathcal{L}}={\left\{l_1,l_2,\ldots,l_N\right\}}$;\\
    \ the adversarial label set ${\mathcal{L}'}={\left\{l'_1,l'_2,\ldots,l'_N\right\}}$;\\
    \ the maximal iterations $M_0$;
}
\Output{
    the adversarial perturbation $\mathbf{r}$;
}
${\mathbf{X}_0}\leftarrow{\mathbf{X}}$, ${\mathbf{r}}\leftarrow{\mathbf{0}}$,
%${M}\leftarrow{0}$, ${\mathcal{T}_0}={\left\{1,2,...,N\right\}}$;\\ ZHISHUAI
${m}\leftarrow{0}$, ${\mathcal{T}_0}\leftarrow\mathcal{T}$;\\ % ZHISHUAI
\While{${m}<{M_0}$ {\bf and} ${\mathcal{T}_m}\neq{\varnothing}$}{
    ${\mathcal{T}_m}={\left\{t_n\mid\arg\max_c\left\{f_c\!\left(\mathbf{X}_m,t_n\right)\right\} = l_n\right\}}$;\\
%    ${\mathbf{r}_m}\leftarrow{{\sum_{n\in\mathcal{T}_m}} ZHISHUAI
    ${\mathbf{r}_m}\leftarrow{{\sum_{t_n\in\mathcal{T}_m}} %ZHISHUAI
        \left[\nabla_{\mathbf{X}_m}f_{l'_n}\!\left(\mathbf{X}_m,t_n\right)-
        \nabla_{\mathbf{X}_m}f_{l_n}\!\left(\mathbf{X}_m,t_n\right)\right]}$;\\
    ${\mathbf{r}'_m}\leftarrow{\frac{\gamma}{\left\|\mathbf{r}_m\right\|_\infty}\mathbf{r}_m}$;\\
    ${\mathbf{r}}\leftarrow{\mathbf{r}+\mathbf{r}'_m}$;\\
%    ${\mathcal{T}_m}={\left\{n\mid\arg\max_c\left\{f_c\!\left(\mathbf{X}_m,t_n\right)\right\} = l_n\right\}}$;\\ ZHISHUAI
    ${\mathbf{X}_{m+1}}\leftarrow{\mathbf{X}_m+\mathbf{r}'_m}$;\\
    ${m}\leftarrow{m+1}$;
}
\Return{
    $\mathbf{r}$
}
\caption{
    Dense Adversary Generation (DAG)
}
\label{Alg:DAG}
\end{algorithm}

We apply a gradient descent algorithm for optimization.
At the $m$-th iteration, denote the current image (possibly after adding several perturbations) as $\mathbf{X}_m$.
We find the set of correctly predicted targets, named the {\em active target set}:
%${\mathcal{T}_m}={\left\{n\mid\arg\max_c\left\{f_c\!\left(\mathbf{X}_m,t_n\right)\right\} = l_n\right\}}$. ZHISHUAI
${\mathcal{T}_m}={\left\{t_n\mid\arg\max_c\left\{f_c\!\left(\mathbf{X}_m,t_n\right)\right\} = l_n\right\}}$. %ZHISHUAI
Then we compute the gradient with respect to the input data and then accumulate all these perturbations:
\begin{equation}
\label{Eqn:Gradient}
\hspace*{-0.08cm}
%{\mathbf{r}_m}={{\sum_{n\in\mathcal{T}_m}}\left[\nabla_{\mathbf{X}_m}f_{l'_n}\!\left(\mathbf{X}_m,t_n\right)- ZHISHUAI
{\mathbf{r}_m}={{\sum_{t_n\in\mathcal{T}_m}}\left[\nabla_{\mathbf{X}_m}f_{l'_n}\!\left(\mathbf{X}_m,t_n\right)- %ZHISHUAI
    \nabla_{\mathbf{X}_m}f_{l_n}\!\left(\mathbf{X}_m,t_n\right)\right]}
\end{equation}
Note that ${\left|\mathcal{T}_m\right|}\ll{\left|\mathcal{T}\right|}$ when $m$ gets large,
thus this strategy considerably reduces the computational overhead.
To avoid numerical instability, we normalize $\mathbf{r}_m$ as
\begin{equation}
\label{Eqn:LearningRate}
{\mathbf{r}'_m}={\frac{\gamma}{\left\|\mathbf{r}_m\right\|_\infty}\cdot\mathbf{r}_m}
\end{equation}
where $\gamma=0.5$ is a fixed hyper-parameter.
We then add $\mathbf{r}'_m$ to the current image $\mathbf{X}_m$ and proceed to the next iteration.
The algorithm terminates if either all the targets are predicted as desired, {\em i.e.}, ${\mathcal{T}_m}={\varnothing}$,
or it reaches the maximum iteration number, which is set to be $200$ in segmentation and $150$ in detection.

The final adversarial perturbation is computed as ${\mathbf{r}}={{\sum_m}\mathbf{r}'_m}$.
Note that, in practice, we often obtain the input image $\mathbf{X}$ after subtracting the mean image $\widehat{\mathbf{X}}$.
In this case, the adversarial image is $\text{Trunc}\!\left(\mathbf{X}+\mathbf{r}+\widehat{\mathbf{X}}\right)$,
where $\text{Trunc}\!\left(\cdot\right)$ denotes the function that truncates every pixel value by $\left[0,255\right]$.
Although truncation may harm the adversarial perturbation, we observed little effect in experiments,
mainly because the magnitude of perturbation $\mathbf{r}$ is very small (see Section~\ref{Generating:Diagnosis:Perceptibility}).
The overall pipeline of DAG algorithm is illustrated in Algorithm~\ref{Alg:DAG}.

\subsection{Selecting Input Proposals for Detection}
\label{Generating:Selection}

A critical issue in DAG is to select a proper set $\mathcal{T}$ of targets.
This is relatively easy in the semantic segmentation task, because the goal is to produce incorrect classification on all pixels,
and thus we can set each of them as a separate target, {\em i.e.}, performing dense sampling on the image lattice.
This is tractable, {\em i.e.}, the computational complexity is proportional to the total number of pixels.

In the scenario of object detection, target selection becomes a lot more difficult,
as the total number of possible targets (bounding box proposals) is orders of magnitudes larger than that in semantic segmentation.
A straightforward choice is to only consider the proposals generated by a sideway network,
{\em e.g.}, the regional proposal network (RPN)~\cite{ren2015faster},
but we find that when the adversarial perturbation $\mathbf{r}$ is added to the original image $\mathbf{X}$,
a different set of proposals may be generated according to the new input $\mathbf{X}+\mathbf{r}$,
and the network may still be able to correctly classify these new proposals~\cite{lou2016foveation}.
To overcome this problem, we make the proposals very dense by increasing the threshold of NMS in RPN.
In practice, when the intersection-over-union (IOU) goes up from $0.70$ to $0.90$,
the average number of proposals on each image increases from around $300$ to around $3000$.
Using this denser target set $\mathcal{T}$,
most probable object bounding boxes are only pixels away from at least one of the selected input proposals,
and we can expect the classification error transfers among neighboring bounding boxes.
As shown in experiments, this heuristic idea works very well,
and the effect of adversarial perturbations is positively correlated to the number of proposals considered in DAG.

Technically, given the proposals generated by RPN, we preserve all {\em positive} proposals and discard the remaining.
Here, a positive proposal satisfies the following two conditions:
1) the IOU with the closest ground-truth object is greater than $0.1$, and
2) the confidence score for the corresponding ground-truth class is greater than $0.1$.
If both conditions hold on multiple ground-truth objects, we select the one with the maximal IOU.
The label of the proposal is defined as the corresponding confident class.
This strategy aims at selecting high-quality targets for Algorithm~\ref{Alg:DAG}.

\renewcommand{\colwidth}{1.0cm}
\begin{table}
\centering
\begin{tabular}{|c||C{\colwidth}|C{\colwidth}|C{\colwidth}|}
\hline
Network           & ORIG    & ADVR    & PERM    \\
\hline\hline
{\bf FCN-Alex}    & $48.04$ & $ 3.98$ & $48.04$ \\
\hline
{\bf FCN-Alex*}     & $48.92$ & $ 3.98$ & $48.91$ \\
\hline
{\bf FCN-VGG}     & $65.49$ & $ 4.09$ & $65.47$ \\
\hline
{\bf FCN-VGG*}     & $67.09$ & $ 4.18$ & $67.08$ \\
\hline\hline
{\bf FR-ZF-07}    & $58.70$ & $ 3.61$ & $58.33$ \\
\hline
{\bf FR-ZF-0712}  & $61.07$ & $ 1.95$ & $60.94$ \\
\hline
{\bf FR-VGG-07}   & $69.14$ & $ 5.92$ & $68.68$ \\
\hline
{\bf FR-VGG-0712} & $72.07$ & $ 3.36$ & $71.97$ \\
\hline
\end{tabular}
\vspace{-0.3cm}
\caption{
    Semantic segmentation (measured by mIOU, $\%$) and object detection (measured by mAP, $\%$) results of different networks.
    Here, ORIG represents the accuracy obtained on the original image set,
    ADVR is obtained on the set after the adversarial perturbations are added,
    and PERM is obtained after the randomly permuted perturbations are added.
    Please see Section~\ref{Generating:Quantitative} for details.
}
\vspace{-0.5cm}
\label{Tab:Quantitative}
\end{table}

\subsection{Quantitative Evaluation}
\label{Generating:Quantitative}

Following some previous work~\cite{szegedy2013intriguing}\cite{moosavi2015deepfool},
we evaluate our approach by measuring the drop in recognition accuracy,
{\em i.e.}, mean intersection-over-union (mIOU) for semantic segmentation and mean average precision (mAP) for object detection,
using the original test images and the ones after adding adversarial perturbations\footnote{For implementation simplicity, we keep  targets with ground-truth class label {\em background} unchanged when generating adversarial examples.}.

\begin{itemize}
\item
For semantic segmentation, we study two network architectures based on the FCN~\cite{long2015fully} framework.
One of them is based on the {\bf AlexNet}~\cite{krizhevsky2012imagenet}
and the other one is based on the $16$-layer {\bf VGGNet}~\cite{simonyan2015very}.
Both networks have two variants.
We use {\bf FCN-Alex} and {\bf FCN-VGG}, which are publicly available,
to denote the networks that are trained on the original FCN~\cite{long2015fully} training set which has $9610$ images,
and use {\bf FCN-Alex*} and {\bf FCN-VGG*} to denote the networks
that are trained on the DeepLab~\cite{chen2016deeplab} training set which has $10582$ images.
We use the validation set in~\cite{long2015fully} which has 736 images as our semantic segmentation test set.
\item
For object detection, based on the Faster-RCNN~\cite{ren2015faster} framework, we study two network architectures,
{\em i.e.}, the {\bf ZFNet}~\cite{zeiler2014visualizing} and the $16$-layer {\bf VGGNet}~\cite{simonyan2015very}.
Both networks have two variants, which are either trained on the {\bf PascalVOC-2007} trainval set,
or the combined {\bf PascalVOC-2007} and {\bf PascalVOC-2012} trainval sets.
These four models are publicly available,
and are denoted as {\bf FR-ZF-07}, {\bf FR-ZF-0712}, {\bf FR-VGG-07} and {\bf FR-VGG-0712}, respectively.
We use the {\bf PascalVOC-2007} test set which has 4952 images as our object detection test set.
\end{itemize}

Results are summarized in Table~\ref{Tab:Quantitative}.
We can observe that the accuracy (mIOU for segmentation and mAP for detection) drops significantly
after the adversarial perturbations are added, demonstrating the effectiveness of DAG algorithm.
Moreover, for detection, the networks with more training data are often more sensitive to the adversarial perturbation.
This is verified by the fact that {\bf FR-ZF-07} (from $58.70\%$ to $3.61\%$)
has a smaller performance drop than {\bf FR-ZF-0712} (from $61.07\%$ to $1.95\%$),
and that {\bf FR-VGG-07} (from $69.14\%$ to $5.92\%$)
has a smaller performance drop than {\bf FR-VGG-0712} (from $72.04\%$ to $3.36\%$).

To verify the importance of the spatial structure of adversarial perturbations,
we evaluate the accuracy after randomly permuting the rows and/or columns of $\mathbf{r}$.
In Table~\ref{Tab:Quantitative}, we find that permuted perturbations cause negligible accuracy drop,
indicating that it is the spatial structure of $\mathbf{r}$,
instead of its magnitude, that indeed contributes in generating adversarial examples.
For permutation results, we randomly permute $\mathbf{r}$ for three times and take the average.

\subsection{Adversarial Examples}
\label{Generating:Examples}

\begin{figure}
\centering
\subfloat{\includegraphics[width=0.48\columnwidth]{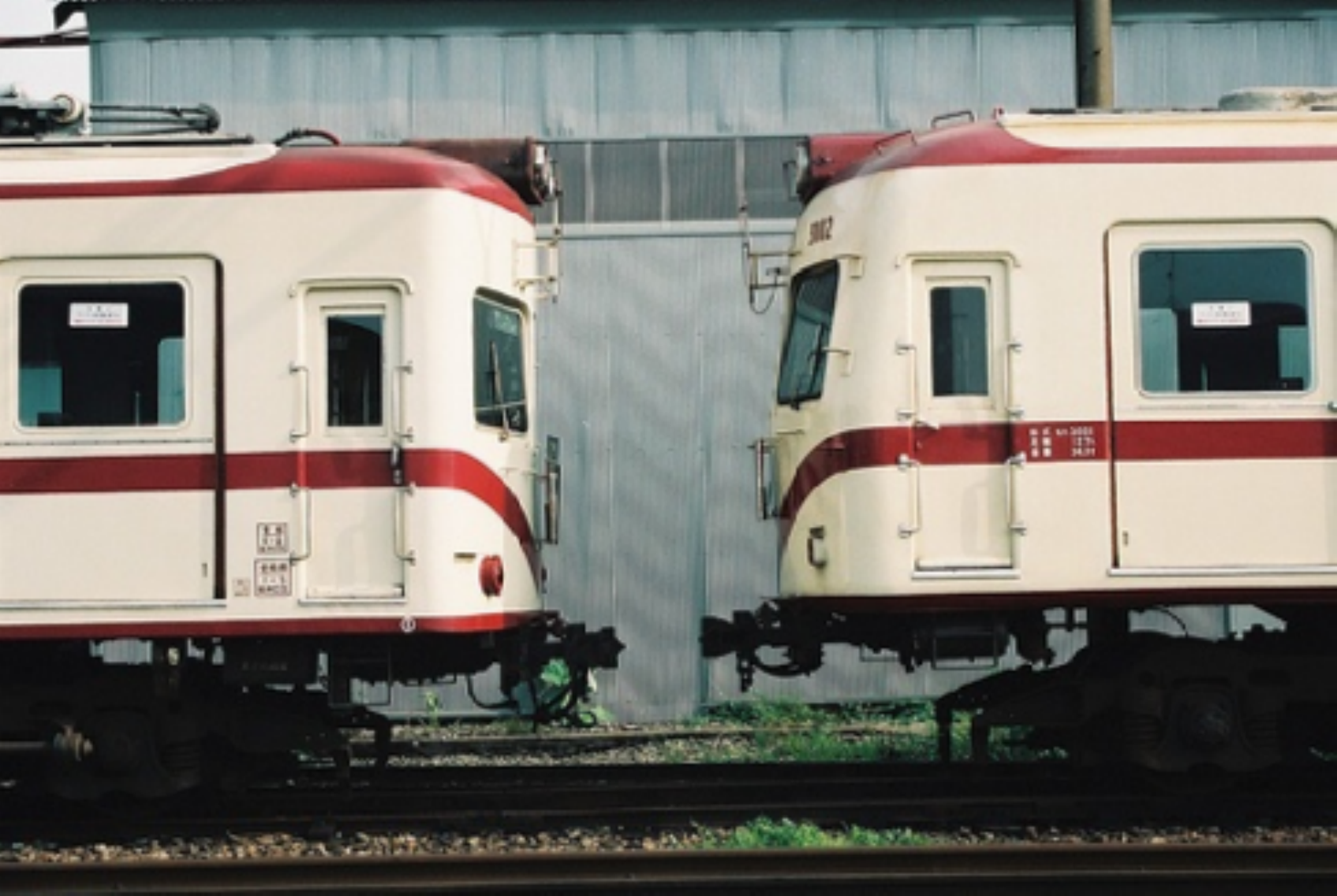}}
\hfill
\subfloat{\includegraphics[width=0.48\columnwidth]{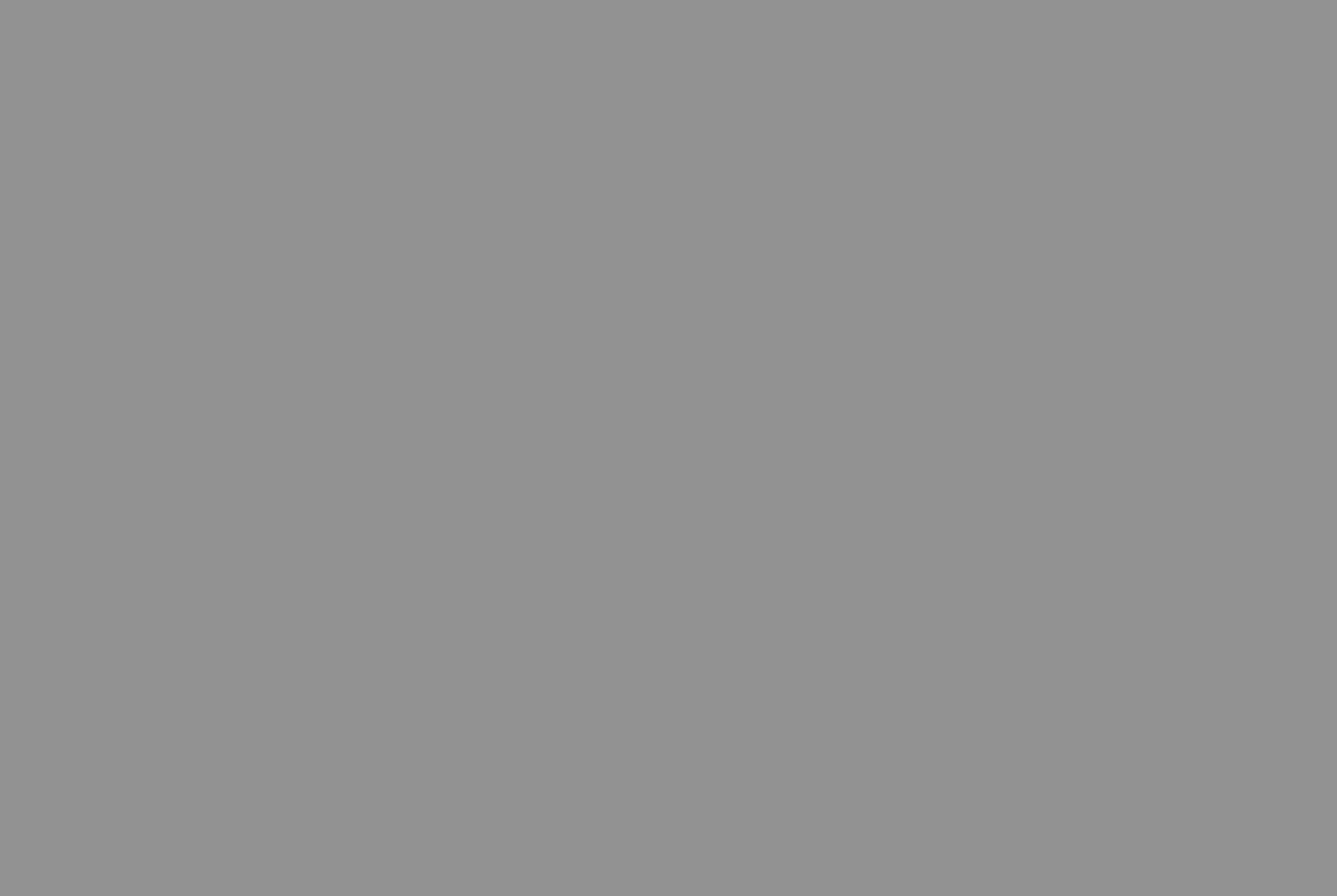}}
\hfill
\subfloat{\includegraphics[width=0.48\columnwidth]{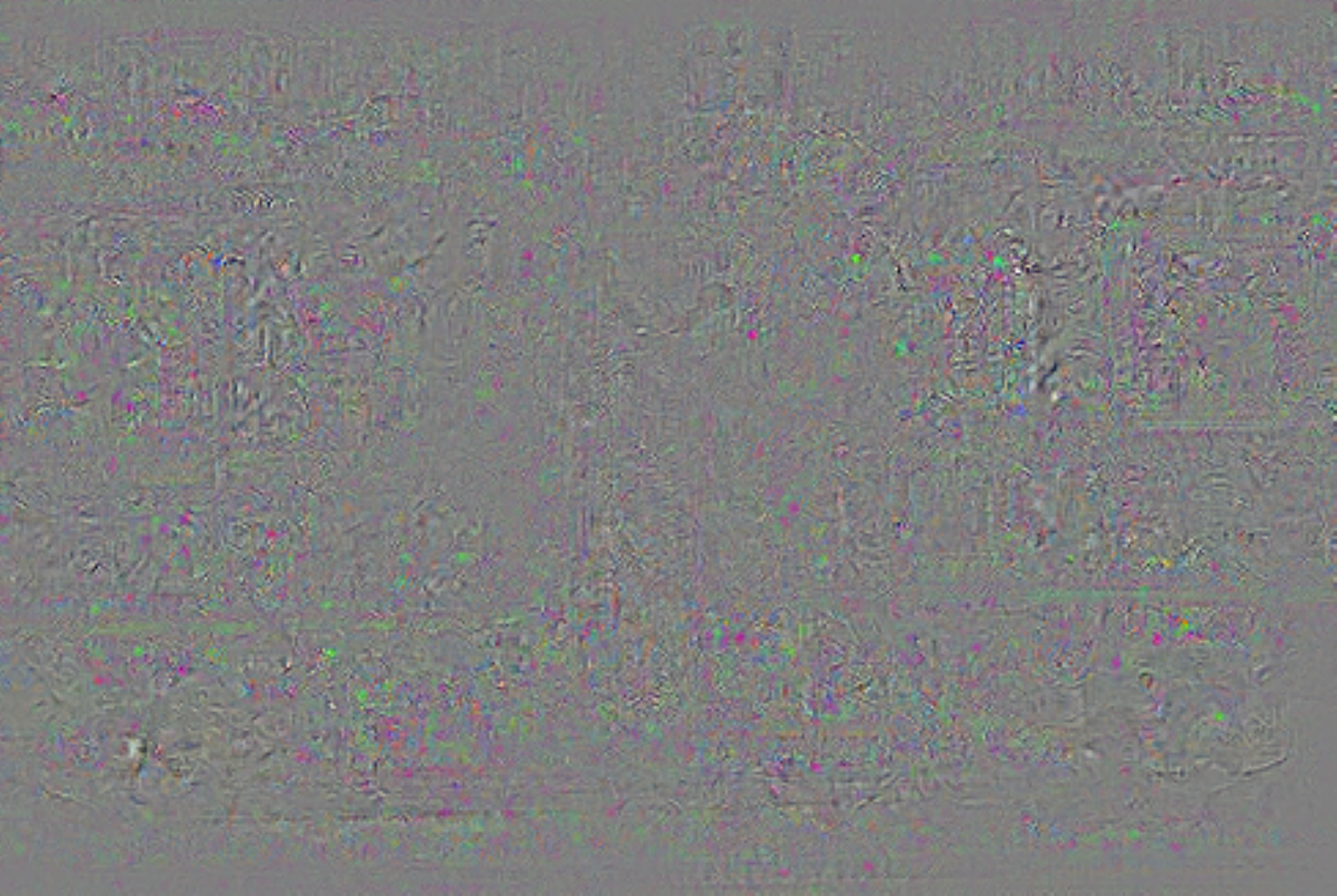}}
\hfill
\subfloat{\includegraphics[width=0.48\columnwidth]{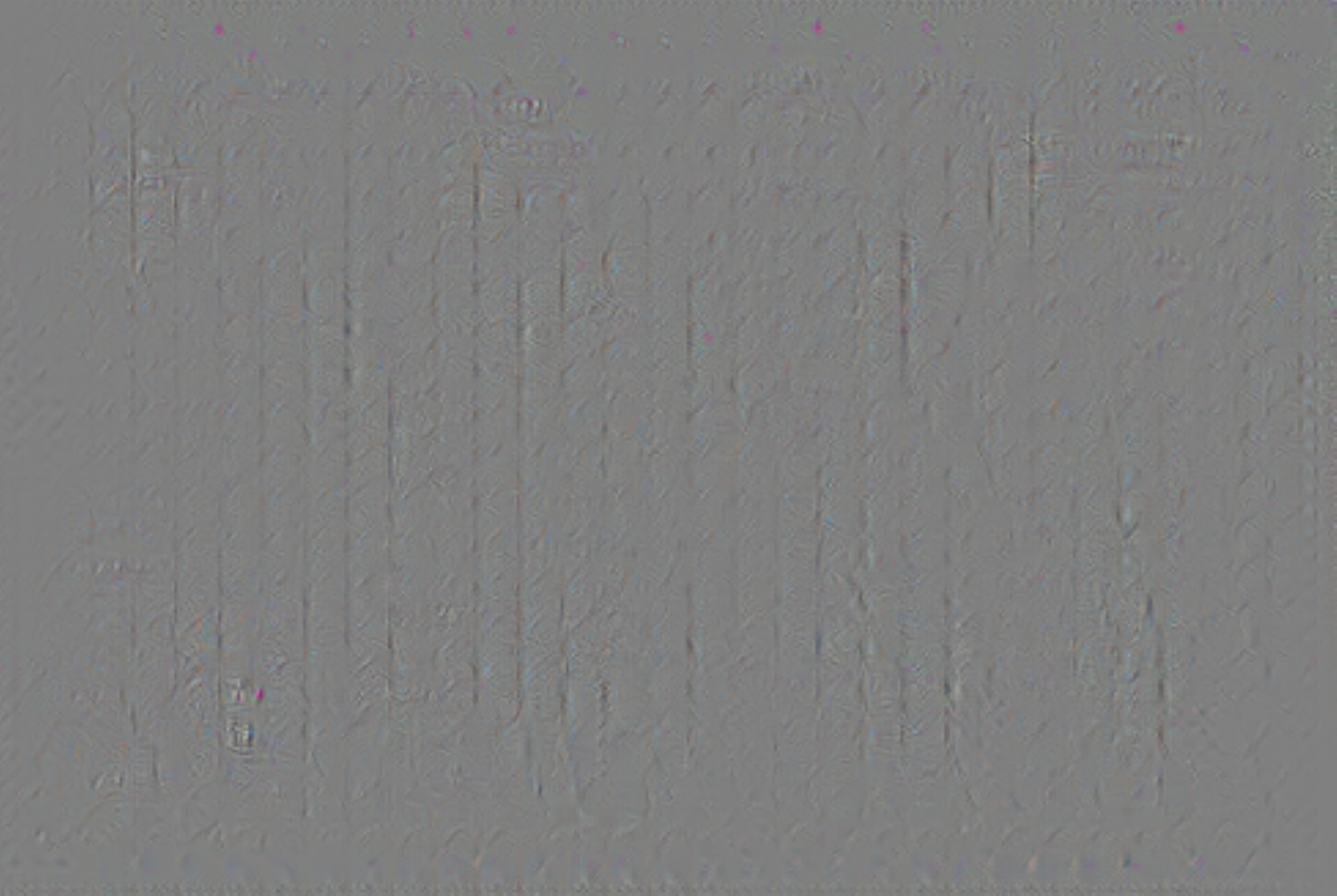}}
\hfill
\subfloat{\includegraphics[width=0.48\columnwidth]{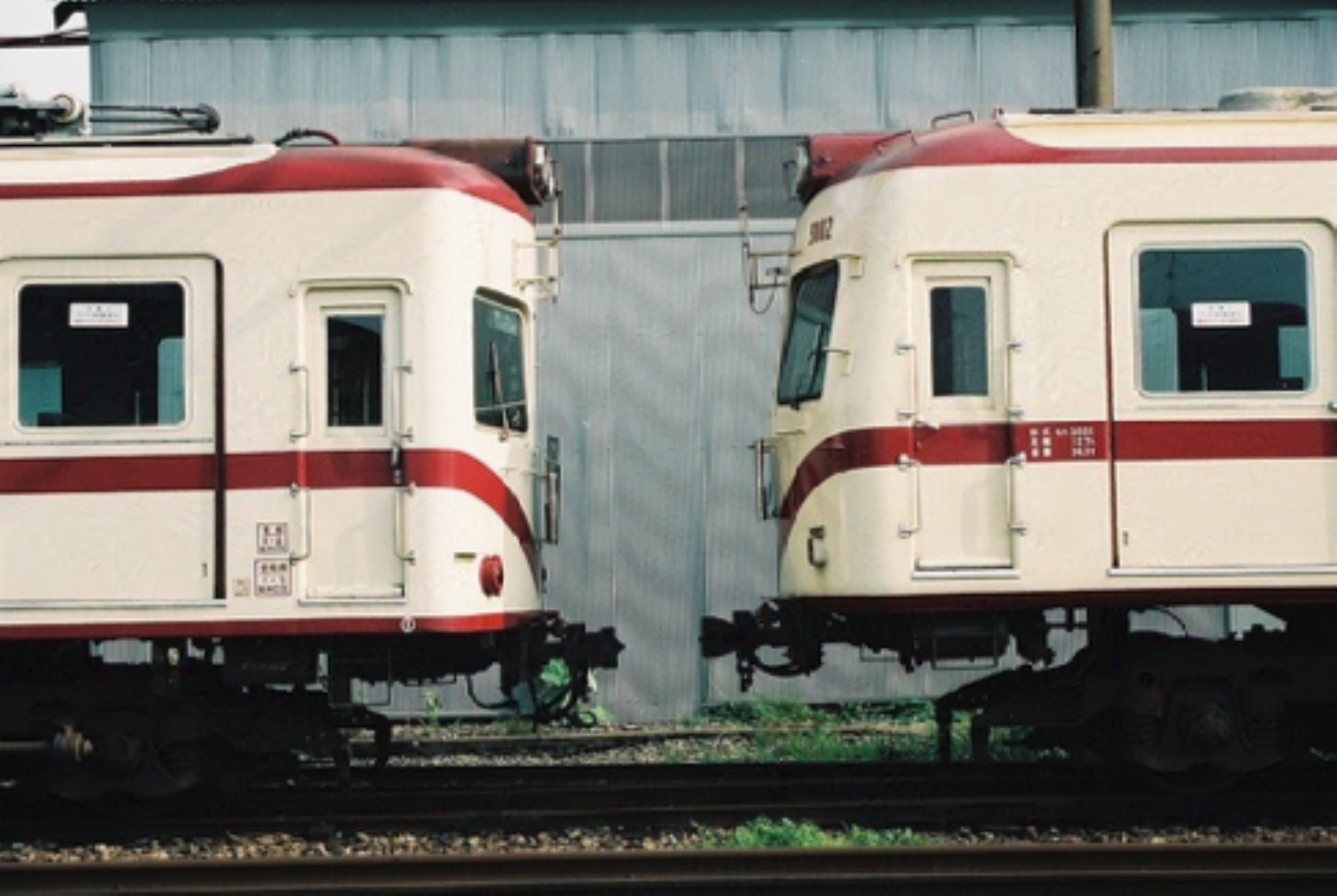}}
\hfill
\subfloat{\includegraphics[width=0.48\columnwidth]{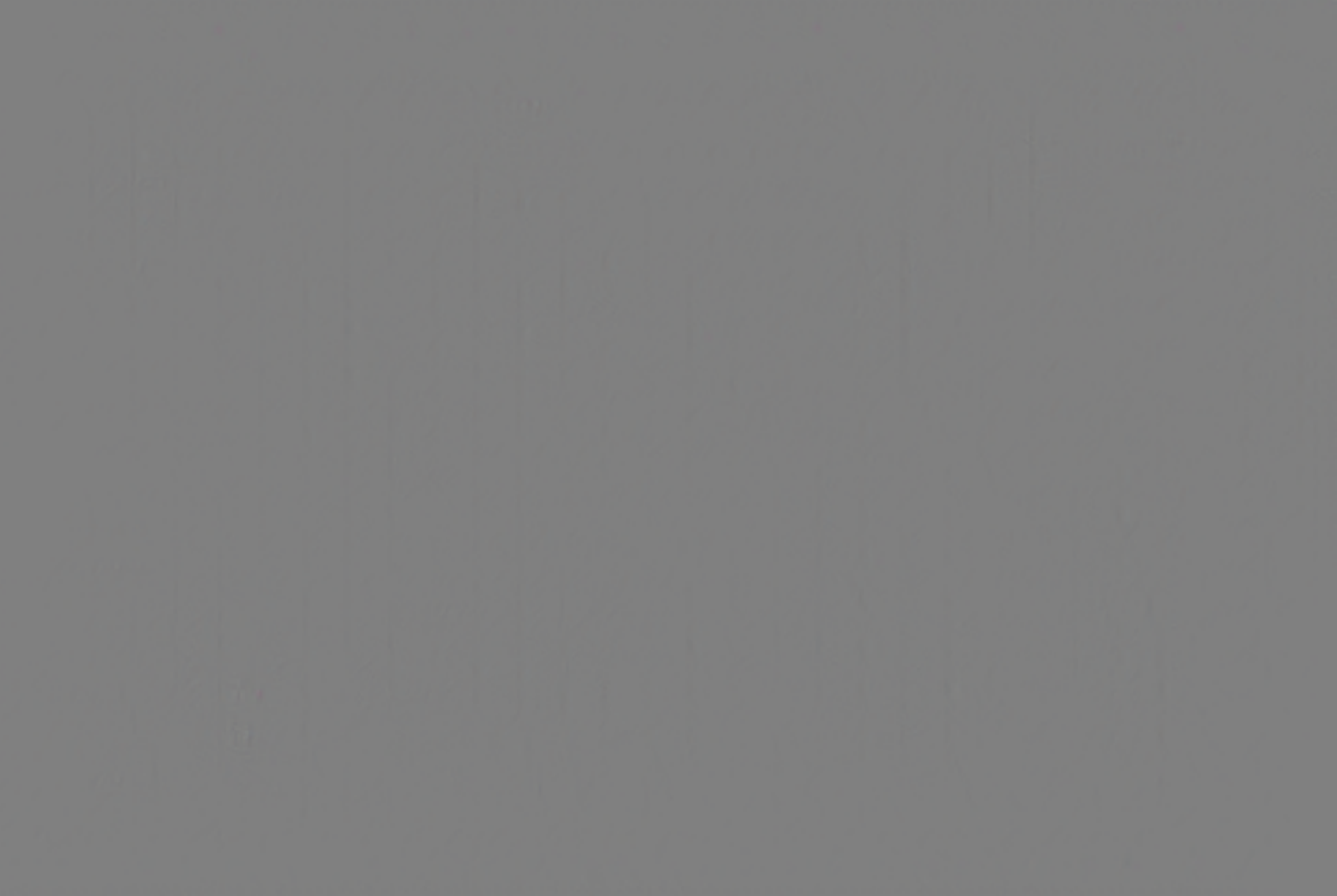}}
\hfill
\subfloat{\includegraphics[width=0.48\columnwidth]{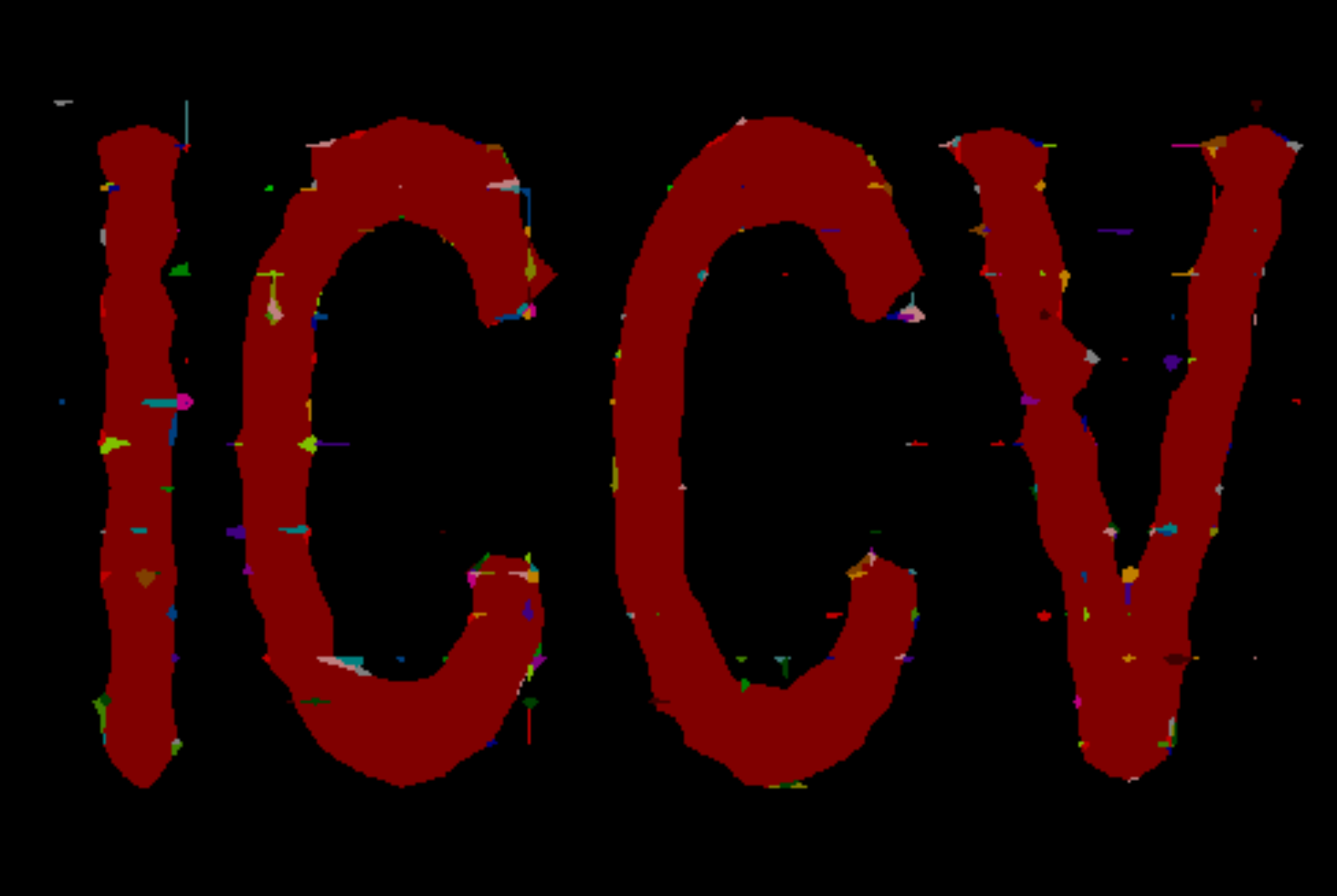}}
\hfill
\subfloat{\includegraphics[width=0.48\columnwidth]{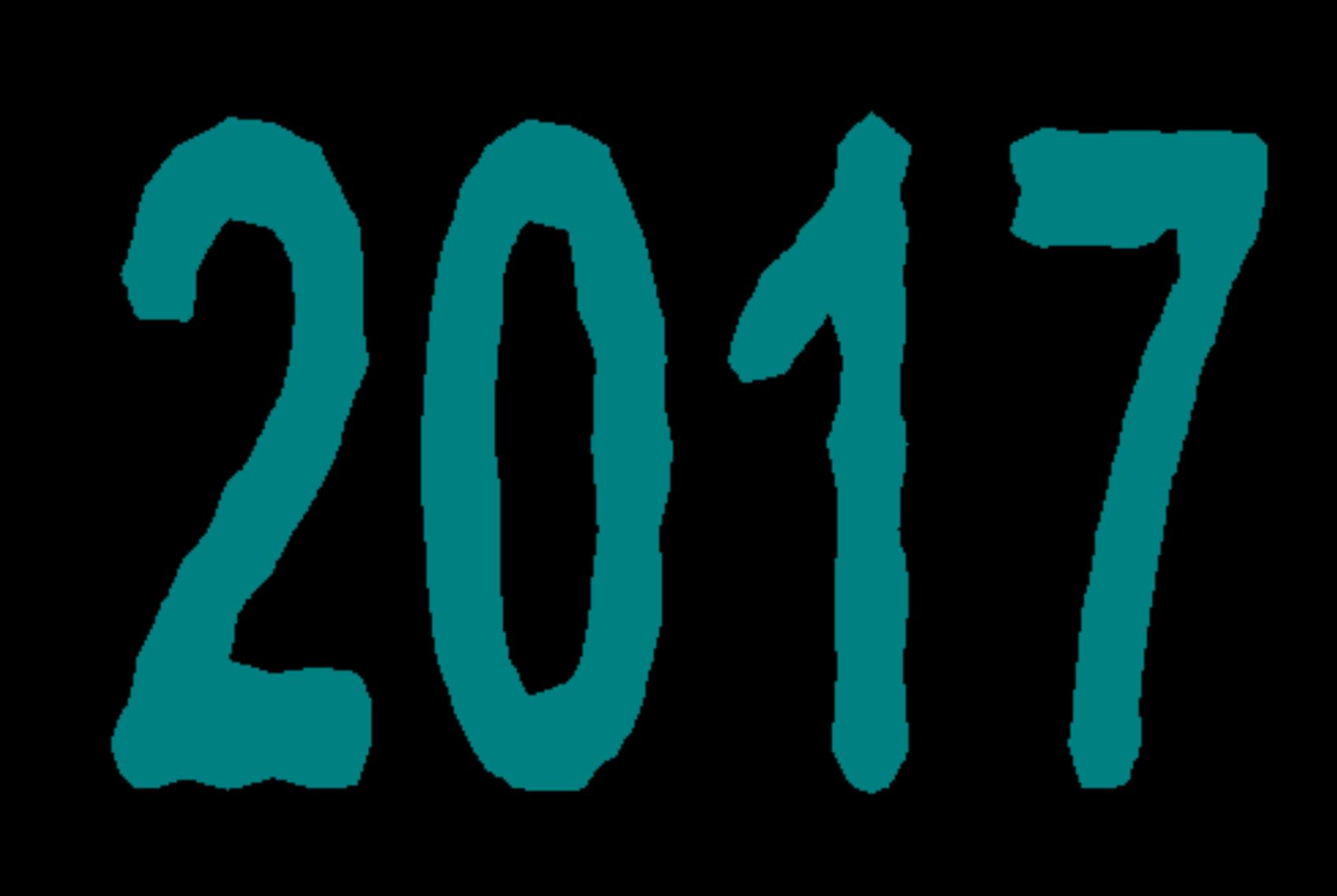}}
\caption{
    Fancy examples generated by DAG for semantic segmentation.
    The {\em adversarial} image is on the left and the {\em fooling} image is on the right.
    From top to bottom: the original image, the perturbation ({\bf magnified by $\mathbf{10}$}),
    the adversarial image after adding perturbation, and the segmentation results.
    The red, blue and black regions are predicted as {\em airplane}, {\em bus} and {\em background}, respectively.
}
\label{Fig:FancyExamples}
\vspace{-0.3cm}
\end{figure}

Figure~\ref{Fig:AdversarialExamples} shows an adversarial example that fails in both detection and segmentation networks.
In addition, we show that DAG is able to control the output of adversarial images very well.
In Figure~\ref{Fig:FancyExamples},
we apply DAG to generating one {\em adversarial} image (which humans can recognize but deep networks cannot)
and one {\em fooling} image~\cite{nguyen2015deep}
(which is completely unrecognizable to humans but deep networks produce false positives).
This suggests that deep networks only cover a limited area in the high-dimensional feature space,
and that we can easily find adversarial and/or fooling examples that fall in the unexplored parts.

\subsection{Diagnostics}
\label{Generating:Diagnosis}

\renewcommand{\scatterwidth}{5.5cm}
\renewcommand{\minipagewidthA}{6.0cm}
\renewcommand{\minipagewidthB}{11.1cm}
\begin{figure*}
\begin{minipage}{\minipagewidthA}
\centering
\includegraphics[width=\scatterwidth]{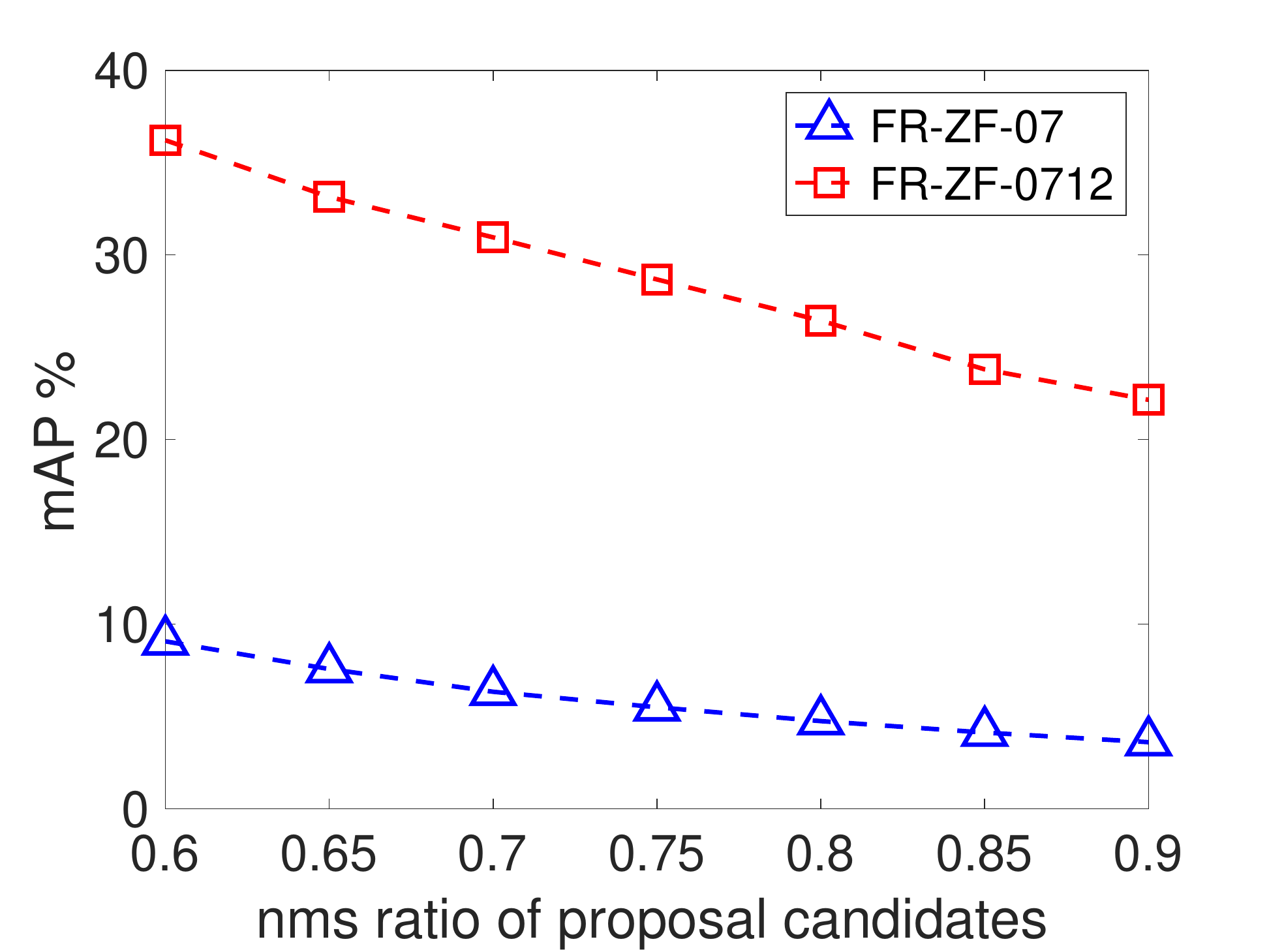}
\caption{
    The mAP of using adversarial perturbations on {\bf FR-ZF-07} to attack {\bf FR-ZF-07} and {\bf FR-ZF-0712},
    with respect to the IOU rate.
    A larger IOU rate leads to a denser set of proposals.
}
\label{Fig:Denseness}
\end{minipage}
\hfill
\begin{minipage}{\minipagewidthB}
\centering
\includegraphics[width=\scatterwidth]{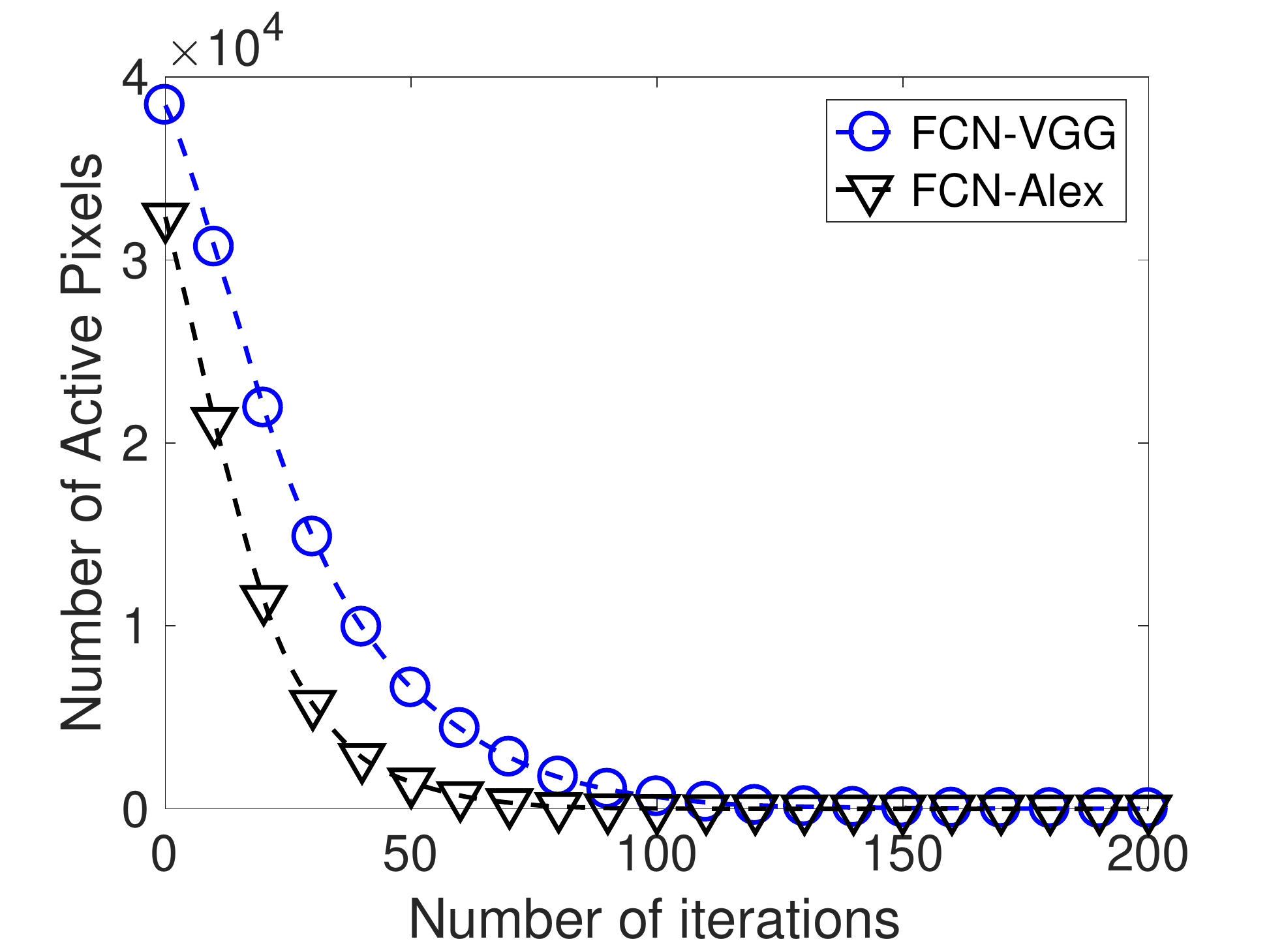}
\includegraphics[width=\scatterwidth]{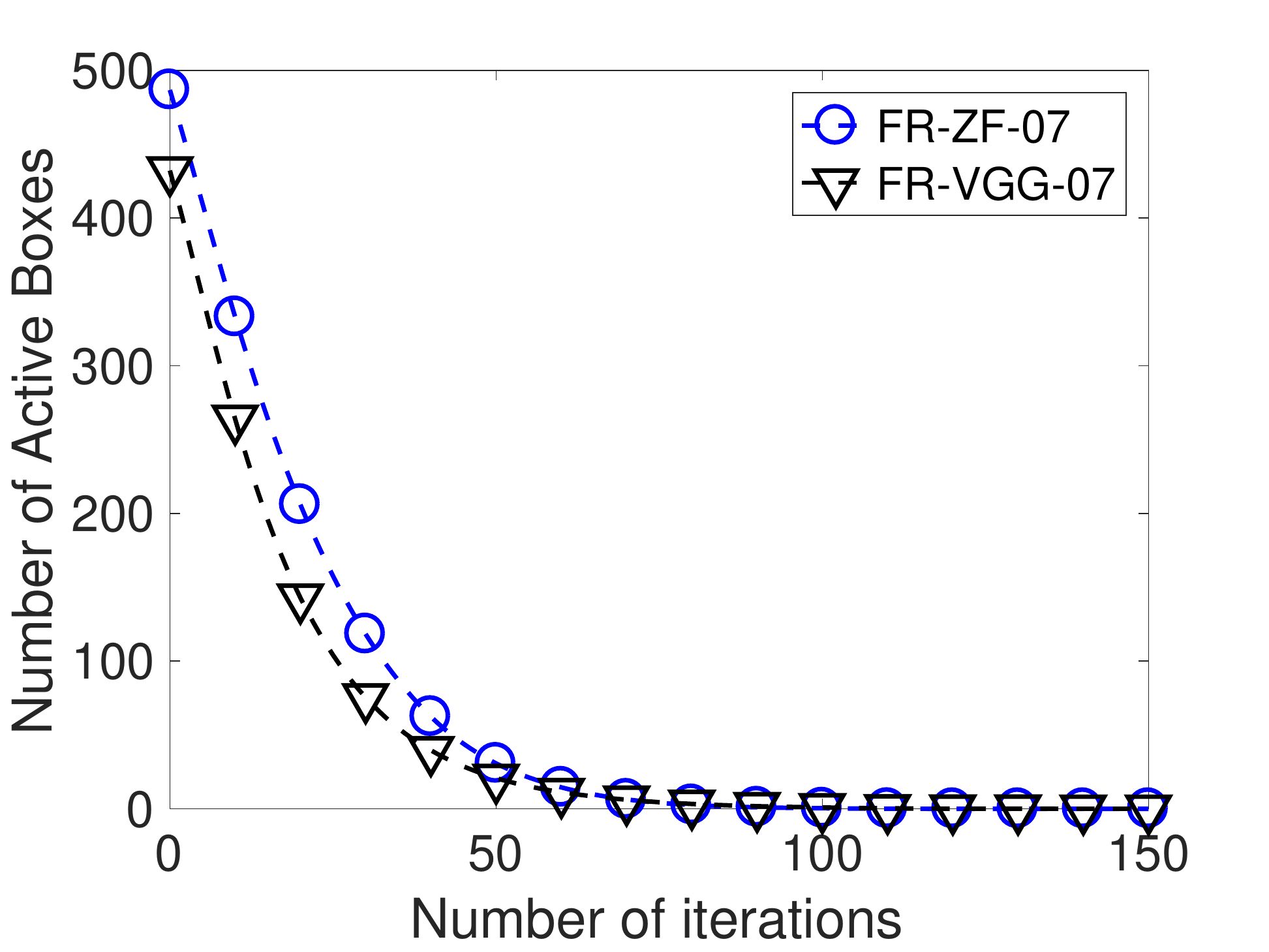}
\caption{
    The convergence of DAG measured by the number of active targets,
    {\em i.e.}, $\left|\mathcal{T}_m\right|$, with respect to the number of iterations.
    Over the entire dataset, the average numbers of iterations are $31.78$ and $54.02$ for {\bf FCN-Alex} and {\bf FCN-VGG},
    and these numbers are $47.05$ and $41.42$ for {\bf FR-ZF-07} and {\bf FR-VGG-07}, respectively.
}
\label{Fig:Convergence}
\end{minipage}
\vspace{-0.2cm}
\end{figure*}

\subsubsection{The Denseness of Proposals}
\label{Generating:Diagnosis:Denseness}

We first observe the impact on adversarial generation of the denseness of the proposals.
To this end, we use different IOU rates in the NMS process after the RPN.
This directly affects the number of proposals preserved in Algorithm~\ref{Alg:DAG}.
As we can see in Figure~\ref{Fig:Denseness},
the mAP value goes down ({\em i.e.}, stronger adversarial perturbations are generated) as the IOU rate increases,
which means that fewer proposals are filtered out and thus the set of targets $\mathcal{T}$ becomes larger.
This is in line of our expectation, since DAG only guarantees misclassification on the targets in $\mathcal{T}$.
The denser sampling on proposals allows the recognition error to propagate to other possible object positions better.
Therefore, we choose a large IOU value ($0.90$) which produces good results.

\subsubsection{Convergence}
\label{Generating:Diagnosis:Convergence}

We then investigate the convergence of DAG,
{\em i.e.}, how many iterations are needed to find the desired adversarial perturbation.
Figure~\ref{Fig:Convergence} shows the number of active targets,
{\em i.e.}, $\left|\mathcal{T}_m\right|$, with respect to the number of iterations $m$.
In general, the training process goes smoothly in the early rounds,
in which we find that the number of active proposals is significantly reduced.
After the algorithm reaches the maximal number of iterations, {\em i.e.}, $200$ in segmentation and $150$ in detection,
only few (less than $1\%$) image fail to converge.
Even on these cases, DAG is able to produce reasonable adversarial perturbations.

Another interesting observation is the difficulty in generating adversarial examples.
In general, the detection networks are more difficult to attack than the segmentation networks,
which is arguably caused by the much larger number of potential targets
(recall that the total number of possible bounding boxes is one or two orders of magnitudes larger).
Meanwhile, as the IOU rate increases, {\em i.e.}, a larger set $\mathcal{T}$ of proposals is considered,
convergence also becomes slower, implying that more iterations are required to generate stronger adversarial perturbations.

\subsubsection{Perceptibility}
\label{Generating:Diagnosis:Perceptibility}

Following~\cite{szegedy2013intriguing}\cite{moosavi2015deepfool},
we compute the perceptibility of the adversarial perturbation $\mathbf{r}$ defined by
${p}={\left(\frac{1}{K}{\sum_k}\left\|\mathbf{r}_k\right\|_2^2\right)^{1/2}}$,
where $K$ is the number of pixels, and $\mathbf{r}_k$ is the intensity vector
($3$-dimensional in the RGB color space, $k=1,2,3$) normalized in $\left[0,1\right]$.
We average the perceptibility value over the entire test set.
In semantic segmentation, these values are $2.6\times10^{-3}$, $2.5\times10^{-3}$, $2.9\times10^{-3}$ and $3.0\times10^{-3}$
on {\bf FCN-Alex}, {\bf FCN-Alex*},  {\bf FCN-VGG} and {\bf FCN-VGG*}, respectively.
In object detection, these values are $2.4\times10^{-3}$, $2.7\times10^{-3}$, $1.5\times10^{-3}$ and $1.7\times10^{-3}$
on {\bf FR-ZF-07}, {\bf FR-ZF-0712}, {\bf FR-VGG-07} and {\bf FR-VGG-0712}, respectively.
One can see that these numbers are very small, which guarantees the imperceptibility of the generated adversarial perturbations.
The visualized examples (Figures~\ref{Fig:AdversarialExamples} and~\ref{Fig:FancyExamples}) also verify this point.

\section{Transferring Adversarial Perturbations}
\label{Transferring}

In this section, we investigate the transferability of the generated adversarial perturbations.
For this respect, we add the adversarial perturbation computed on one model to attack other models.
The attacked model may be trained based on a different (sometimes unknown) network architecture,
or even targeted at a different vision task.
Quantitative results are summarized in Tables~\ref{Tab:TransferDet} -~\ref{Tab:TransferDetSeg},
and typical examples are illustrated in Figure~\ref{Fig:TransferrableExamples}.
In the following parts, we analyze these results by organizing them into three categories,
namely {\em cross-training} transfer, {\em cross-network} transfer and {\em cross-task} transfer.

\renewcommand{\colwidth}{1.89cm}%zhishuai
\begin{table*}
\centering
\begin{tabular}{|C{3cm}||C{\colwidth}|C{\colwidth}|C{\colwidth}|C{\colwidth}|C{\colwidth}|C{\colwidth}|}
\hline
Adversarial Perturbations from           & {{\bf FR-ZF-07}}
                                         & {{\bf FR-ZF-0712}}
                                         & {{\bf FR-VGG-07}}
                                         & {{\bf FR-VGG-0712}}
                                         & {{\bf R-FCN-RN50}}
                                         & {{\bf R-FCN-RN101}}
\\
\hline
{\bf None}                               & {$58.70$}              & {$61.07$}
                                         & {$69.14$}              & {$72.07$}
                                         & {$76.40$}              & {$78.06$}
\\
\hline
{{\bf FR-ZF-07} ($\mathbf{r}_1$)}
                                         & {$ \mathbf{3.61}$}              & {$22.15$}
                                         & {$66.01$}              & {$69.47$} %zhishuai
                                         & {$74.01$}              & {$75.87$}
\\
\hline
{{\bf FR-ZF-0712} ($\mathbf{r}_2$)}
                                         & {$13.14$}              & {$ \mathbf{1.95}$}
                                         & {$64.61$}              & {$68.17$}
                                         & {$72.29$}              & {$74.68$}
\\
\hline
{{\bf FR-VGG-07} ($\mathbf{r}_3$)}
                                         & {$56.41$}              & {$59.31$}
                                         & {$ \mathbf{5.92}$}              & {$48.05$}
                                         & {$72.84$}              & {$74.79$} %zhishuai
\\
\hline
{{\bf FR-VGG-0712} ($\mathbf{r}_4$)}
                                         & {$56.09$}              & {$58.58$}
                                         & {$31.84$}              & {$ \mathbf{3.36}$}
                                         & {$70.55$}              & {$72.78$}
\\
\hline
{$\mathbf{r}_1+\mathbf{r}_3$}
                                         & {$ \mathbf{3.98}$}              & {$21.63$}
                                         & {$ \mathbf{7.00}$}              & {$44.14$}
                                         & {$68.89$}              & {$71.56$}
\\
\hline
{$\mathbf{r}_1+\mathbf{r}_3 \ \text{(permute)}$}
                                         & {$58.30$}              & {$61.08$}%zhishuai
                                         & {$68.63$}              & {$71.82$}%zhishuai
                                         & {$76.34$}              & {$77.71$}%zhishuai
\\
\hline
{$\mathbf{r}_2+\mathbf{r}_4$}
                                         & {$13.15$}              & {$ \mathbf{2.13}$}
                                         & {$28.92$}              & {$ \mathbf{4.28}$}
                                         & {$63.93$}              & {$67.25$}
\\
\hline
{$\mathbf{r}_2+\mathbf{r}_4 \ \text{(permute)}$}
                                         & {$58.51$}              & {$ 61.09$}%zhishuai
                                         & {$68.68$}              & {$ 71.78$}%zhishuai
                                         & {$76.23$}              & {$77.71$}%zhishuai
\\
\hline
\end{tabular}
\vspace{-0.3cm}
\caption{
    Transfer results for detection networks.
    {\bf FR-ZF-07}, {\bf FR-ZF-0712}, {\bf FR-VGG-07} and {\bf FR-VGG-0712}
    are used as four basic models to generate adversarial perturbations,
    and {\bf R-FCN-RN50} and {\bf R-FCN-RN101} are used as black-box models.
    All models are evaluated on the {\bf PascalVOC-2007} test set and its adversarial version, which both has $4952$ images.
}
\label{Tab:TransferDet}
\vspace{-0.15cm}
\end{table*}

\renewcommand{\colwidth}{1.89cm}%zhishuai
\begin{table*}
\centering
\begin{tabular}{|C{3cm}||C{\colwidth}|C{\colwidth}|C{\colwidth}|C{\colwidth}|C{\colwidth}|C{\colwidth}|}
\hline
Adversarial Perturbations from           & {{\bf FCN-Alex}}
                                         & {{\bf FCN-Alex*}}
                                         & {{\bf FCN-VGG}}
                                         & {{\bf FCN-VGG*}}
                                         & {{\bf DL-VGG}}
                                         & {{\bf DL-RN101}}
 \\
\hline
{\bf None}                               & {$48.04$}              & {$48.92$}
                                         & {$65.49$}              & {$67.09$}
                                         & {$70.72$}              & {$76.11$}
 \\
\hline
{{\bf FCN-Alex} ($\mathbf{r}_5$)}
                                         & {$ \mathbf{3.98}$}              & {$7.94$}
                                         & {$64.82$}              & {$66.54$}
                                         & {$70.18$}              & {$75.45$}
\\
\hline
{{\bf FCN-Alex*} ($\mathbf{r}_6$)}
                                         & {$ 5.10$}              & {$ \mathbf{3.98}$}
                                         & {$64.60$}              & {$66.36$}
                                         & {$69.98$}              & {$75.52$}
 \\
\hline
{{\bf FCN-VGG} ($\mathbf{r}_7$)}
                                         & {$46.21$}              & {$47.38$}
                                         & {$ \mathbf{4.09}$}              & {$16.36$}
                                         & {$45.16$}              & {$73.98$}
 \\
\hline
{{\bf FCN-VGG*} ($\mathbf{r}_8$)}
                                         & {$46.10$}              & {$47.21$}
                                         & {$12.72$}              & {$ \mathbf{4.18}$}
                                         & {$46.33$}              & {$73.76$}
 \\
\hline
{$\mathbf{r}_5+\mathbf{r}_7$} %zhishuai
                                         & {$ \mathbf{4.83}$}              & {$ 8.55$}
                                         & {$ \mathbf{4.23}$}              & {$17.59$}
                                         & {$43.95$}              & {$73.26$}
 \\
\hline
{$\mathbf{r}_5+\mathbf{r}_7 \ \text{(permute)}$} %zhishuai
                                         & {$48.03$}              & {$48.90$}
                                         & {$65.47$}              & {$67.09$}
                                         & {$70.69$}              & {$76.04$}
 \\
\hline
{$\mathbf{r}_6+\mathbf{r}_8$} %zhishuai
                                         & {$ 5.52$}              & {$ \mathbf{4.23}$}
                                         & {$13.89$}              & {$ \mathbf{4.98}$}
                                         & {$44.18$}              & {$73.01$}
 \\
\hline
{$\mathbf{r}_6+\mathbf{r}_8 \ \text{(permute)}$} %zhishuai
                                         & {$48.03$}              & {$48.90$}
                                         & {$65.47$}              & {$67.05$}
                                         & {$70.69$}              & {$76.05$}
 \\
\hline
\end{tabular}
\vspace{-0.3cm}
\caption{
    Transfer results for segmentation networks.
    {\bf FCN-Alex}, {\bf FCN-Alex*}, {\bf FCN-VGG} and {\bf FCN-VGG*}
    are used as four basic models to generate adversarial perturbations,
    and {\bf DL-VGG} and {\bf DL-RN101} are used as black-box models.
    All models are evaluated on validation set in \cite{long2015fully} and its adversarial version, which both has $736$ images.
}
\label{Tab:TransferSeg}
\vspace{-0.15cm}
\end{table*}

\renewcommand{\colwidth}{2.354cm}%zhishuai
\begin{table*}[htpb!]
\centering
\begin{tabular}{|C{3cm}||C{\colwidth}|C{\colwidth}|C{\colwidth}|C{\colwidth}|C{\colwidth}|}
\hline
Adversarial Perturbations from           & {{\bf FR-ZF-07}}
                                         & {{\bf FR-VGG-07}}
                                         & {{\bf FCN-Alex}}
                                         & {{\bf FCN-VGG}}
                                         & {{\bf R-FCN-RN101}}
\\
\hline
{\bf None}                   & {$56.83$}
                                         & {$68.88$}
                                         & {$35.73$}
                                         & {$54.87$}           & {$80.20$}
                                                   \\
\hline
{{\bf FR-ZF-07} ($\mathbf{r}_1$)} %new
                                         & {$\mathbf{5.14}$}
                                         & {$ 66.63$}
                                         & {$31.74$}
                                         & {$51.94$}           & {$76.00$}
                                                    \\
                                         \hline
{{\bf FR-VGG-07} ($\mathbf{r}_3$)} %new
                                         & {$54.96$}
                                         & {$ \mathbf{7.17}$}
                                         & {$34.53$}
                                         & {$43.06$}           & {$74.50$}
                                                    \\
                                         \hline

{{\bf FCN-Alex} ($\mathbf{r}_5$)} %new
                                         & {$55.61$}
                                         & {$ 68.62$}
                                         & {$\mathbf{4.04}$}
                                         & {$54.08$}           & {$77.09$}
                                                    \\
                                         \hline
{{\bf FCN-VGG} ($\mathbf{r}_7$)} %new
                                         & {$55.24$}
                                         & {$56.33$}
                                         & {$33.99$}
                                         & {$\mathbf{4.10}$}           & {$73.86$}
                                                   \\
\hline
{$\mathbf{r}_1+\mathbf{r}_3+\mathbf{r}_5$}
                                         & {$ \mathbf{5.02}$}
                                         & {$ \mathbf{8.75}$}
                                         & {$ \mathbf{4.32}$}
                                         & {$ 37.90$}           & {$69.07$}
                                                   \\
\hline
{$\mathbf{r}_1+\mathbf{r}_3+\mathbf{r}_7$}
                                         & {$ \mathbf{5.15}$}
                                         & {$ \mathbf{5.63}$}
                                         & {$28.48$}
                                         & {$\mathbf{4.81}$}           & {$ 65.23$}
                                                   \\
\hline
{$\mathbf{r}_1+\mathbf{r}_5+\mathbf{r}_7$}
                                         & {$ \mathbf{5.14}$}
                                         & {$47.52$}
                                         & {$\mathbf{4.37}$}
                                         & {$ \mathbf{5.20}$}           & {$ 68.51$}
                                                   \\
\hline
{$\mathbf{r}_3+\mathbf{r}_5+\mathbf{r}_7$}
                                         & {$53.34$}
                                         & {$ \mathbf{5.94}$}
                                         & {$\mathbf{4.41}$}
                                         & {$ \mathbf{4.68}$}           & {$ 67.57$}
                                                    \\
\hline
{$\mathbf{r}_1+\mathbf{r}_3+\mathbf{r}_5+\mathbf{r}_7$}
                                         & {$ \mathbf{5.05}$}
                                         & {$ \mathbf{5.89}$}
                                         & {$ \mathbf{4.51}$}
                                         & {$ \mathbf{5.09}$}           & {$ 64.52$}
                                                    \\
\hline
\end{tabular}
\vspace{-0.3cm}
\caption{
    Transfer results between detection networks and segmentation networks.
    {\bf FR-ZF-07}, {\bf FR-VGG-07}, {\bf FCN-Alex} and {\bf FCN-VGG}
    are used as four basic models to generate adversarial perturbations, and {\bf R-FCN-RN101} are used as black-box model.
    When attacking the first four basic networks,
    we use a subset of the {\bf PascalVOC-2012} segmentation validation set which contains $687$ images.
    %In the black-box attack, {\em i.e.}, we only evaluate our method on the non-intersecting subset of $110$ images. ZHISHUAI
    In the black-box attack, we evaluate our method on the non-intersecting subset of $110$ images. %ZHISHUAI
}
\label{Tab:TransferDetSeg}
\vspace{-0.3cm}
\end{table*}

\subsection{Cross-Training Transfer}
\label{Transferring:CrossTraining}

By {\em cross-training} transfer, we mean to apply the perturbations learned from one network
to another network with the same architecture but trained on a different dataset.
We observe that the transferability {\em largely} exists within the same network structure\footnote{We also studied training on strictly non-overlapping datasets, e.g., the model {\bf FR-ZF-07} trained on {\bf PascalVOC-2007} trainval set and the model {\bf FR-ZF-12val} trained on {\bf PascalVOC-2012} val set. The experiments deliver similar conclusions. For example, using {\bf FR-ZF-07} to attack {\bf FR-ZF-12val} results in a mAP drop from $56.03\%$ to $25.40\%$, and using {\bf FR-ZF-12val} to attack {\bf FR-ZF-07} results in a mAP drop from $58.70\%$ to $30.41\%$.}.
For example, using the adversarial perturbations generated by {\bf FR-ZF-07} to attack {\bf FR-ZF-0712} obtains a $22.15\%$ mAP.
This is a dramatic drop from the performance ($61.07\%$) reported on the original images,
although the drop is less than that observed in attacking {\bf FR-ZF-07} itself (from $58.70\%$ to $3.61\%$).
Meanwhile, using the adversarial perturbations generated by {\bf FR-ZF-0712} to attack {\bf FR-ZF-07}
causes the mAP drop from $58.70\%$ to $13.14\%$,
We observe similar phenomena when {\bf FR-VGG-07} and {\bf FR-VGG-0712}, or {\bf FCN-Alex} and {\bf FCN-Alex*},
or {\bf FCN-VGG} and {\bf FCN-VGG*} are used to attack each other.
Detailed results are shown in Tables~\ref{Tab:TransferDet} and \ref{Tab:TransferSeg}.

\subsection{Cross-Network Transfer}
\label{Transferring:CrossNetwork}

We extend the previous case to consider the transferability through different network structures.
We introduce two models which are more powerful than what we used to generate adversarial perturbations,
namely DeepLab~\cite{chen2016deeplab} for semantic segmentation and R-FCN~\cite{li2016r} for object detection.
For DeepLab~\cite{chen2016deeplab}, we use {\bf DL-VGG} to denote the network based on 16-layer {\bf VGGNet}\cite{simonyan2015very},
and use {\bf DL-RN101} to denote the network based on 101-layer {\bf ResNet}\cite{he2016deep}.
Both networks are trained on original DeepLab~\cite{chen2016deeplab} training set which has $10582$ images.
For R-FCN~\cite{li2016r}, we use {\bf R-FCN-RN50} to denote the network based on 50-layer {\bf ResNet}\cite{he2016deep},
and use {\bf R-FCN-RN101} to denote the network based on 101-layer {\bf ResNet}\cite{he2016deep}.
Both networks are trained on the combined trainval sets of {\bf PascalVOC-2007} and {\bf PascalVOC-2012}.
The perturbations applied to these four models are considered as black-box attacks~\cite{papernot2016practical},
since DAG does not know the structure of these networks beforehand.

Detailed results are shown in Tables~\ref{Tab:TransferDet} and~\ref{Tab:TransferSeg}.
Experiments reveal that transferability between different network structures becomes weaker.
For example, applying the perturbations generated by {\bf FR-ZF-07} leads to slight accuracy drop
on {\bf FR-VGG-07} (from $69.14\%$ to $66.01\%$), {\bf FR-VGG-0712} (from $72.07\%$ to $69.74\%$),
{\bf R-FCN-RN50} (from $76.40\%$ to $74.01\%$) and {\bf R-FCN-RN101} (from $78.06\%$ to $75.87\%$), respectively.
Similar phenomena are observed in using different segmentation models to attack each other.
One exception is using {\bf FCN-VGG} or {\bf FCN-VGG*} to attack {\bf DL-VGG}
(from $70.72\%$ to $45.16\%$ for {\bf FCN-VGG} attack, or from $70.72\%$ to $46.33\%$ by {\bf FCN-VGG*} attack),
which results in a significant accuracy drop of {\bf DL-VGG}.
Considering the cues obtained from previous experiments,
we conclude that adversarial perturbations are closely related to the architecture of the network.

\subsection{Cross-Task Transfer}
\label{Transferring:CrossTask}

Finally, we investigate {\em cross-task} transfer,
{\em i.e.}, using the perturbations generated by a detection network to attack a segmentation network or in the opposite direction.
We use a subset of {\bf PascalVOC-2012} segmentation validation set as our test set
\footnote{There are training images of {\bf FR-ZF-07}, {\bf FR-VGG-07}, {\bf FCN-Alex} and {\bf FCN-VGG}
included in the {\bf PascalVOC-2012} segmentation validation set, so we validate on the non-intersecting set of $687$ images.}.
Results are summarized in Table~\ref{Tab:TransferDetSeg}.
We note that if the same network structure is used,
{\em e.g.}, using {\bf FCN-VGG} (segmentation) and {\bf FR-VGG-07} (detection) to attack each other,
the accuracy drop is significant (the mIOU of {\bf FCN-VGG} drops from $54.87\%$ to $43.06\%$,
and the mAP of {\bf FR-VGG-07} drops from $68.88\%$ to $56.33\%$).
Note that this drop is even more significant than {\em cross-network} transfer on the same task,
which verifies our hypothesis again that the adversarial perturbations are related to the network architecture.

\subsection{Combining Heterogeneous Perturbations}
\label{Transferring:Combination}

From the above experiments, we assume that different network structures generate roughly {\em orthogonal} perturbations,
which means that if $\mathbf{r}_\mathbb{A}$ is generated by one structure $\mathbb{A}$,
then adding it to another structure $\mathbb{B}$ merely changes the recognition results,
{\em i.e.}, ${\mathbf{f}^\mathbb{B}\!\left(\mathbf{X},t_n\right)}\approx
    {\mathbf{f}^\mathbb{B}\!\left(\mathbf{X}+\mathbf{r}_\mathbb{A},t_n\right)}$.
This motivates us to combine heterogeneous perturbations towards better adversarial performance.
For example, if both $\mathbf{r}_\mathbb{A}$ and $\mathbf{r}_\mathbb{B}$ are added,
we have ${\mathbf{f}^\mathbb{A}\!\left(\mathbf{X}+\mathbf{r}_\mathbb{A}+\mathbf{r}_\mathbb{B},t_n\right)}\approx
    {\mathbf{f}^\mathbb{A}\!\left(\mathbf{X}+\mathbf{r}_\mathbb{A},t_n\right)}$ and
${\mathbf{f}^\mathbb{B}\!\left(\mathbf{X}+\mathbf{r}_\mathbb{A}+\mathbf{r}_\mathbb{B},t_n\right)}\approx
    {\mathbf{f}^\mathbb{B}\!\left(\mathbf{X}+\mathbf{r}_\mathbb{B},t_n\right)}$.
Thus, the combined perturbation $\mathbf{r}_\mathbb{A}+\mathbf{r}_\mathbb{B}$ is able to confuse both network structures.

\begin{figure}[!htb]
\centering
\includegraphics[width=\columnwidth]{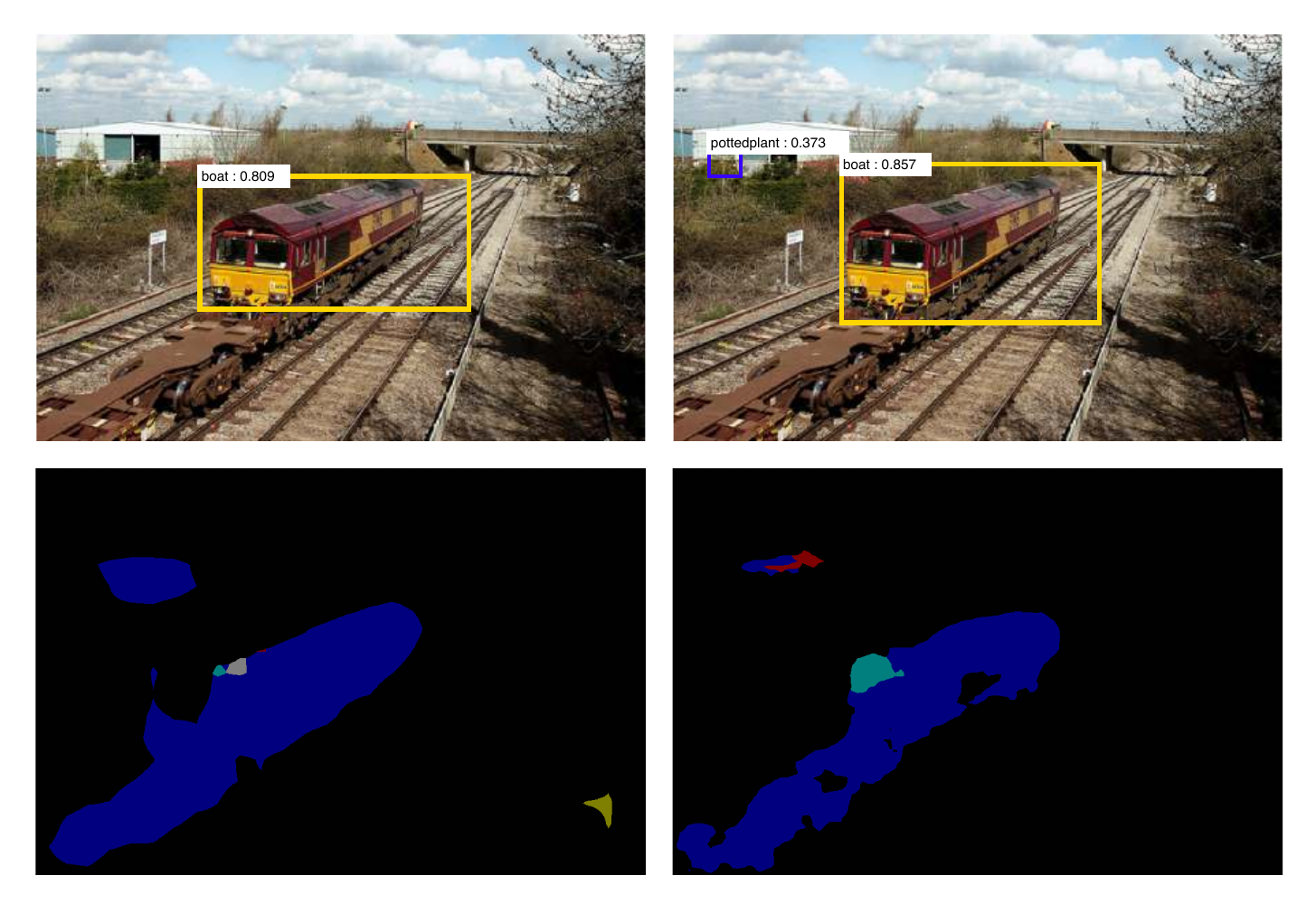}
\vspace{-0.7cm}
\caption{
    Adding one adversarial perturbation computed by $\mathbf{r}_1+\mathbf{r}_3+\mathbf{r}_5+\mathbf{r}_7$
    (see Table~\ref{Tab:TransferDetSeg}) confuses four different networks.
    The top row shows {\bf FR-VGG-07} and {\bf FR-ZF-07} detection results,
    and the bottom row shows {\bf FCN-Alex} and {\bf FCN-VGG} segmentation results.
    The blue in segmentation results corresponds to boat.
}
\vspace{-0.3cm}
\label{Fig:FoolAll}
\end{figure}

\begin{figure*}
\centering
\includegraphics[width=0.95\linewidth]{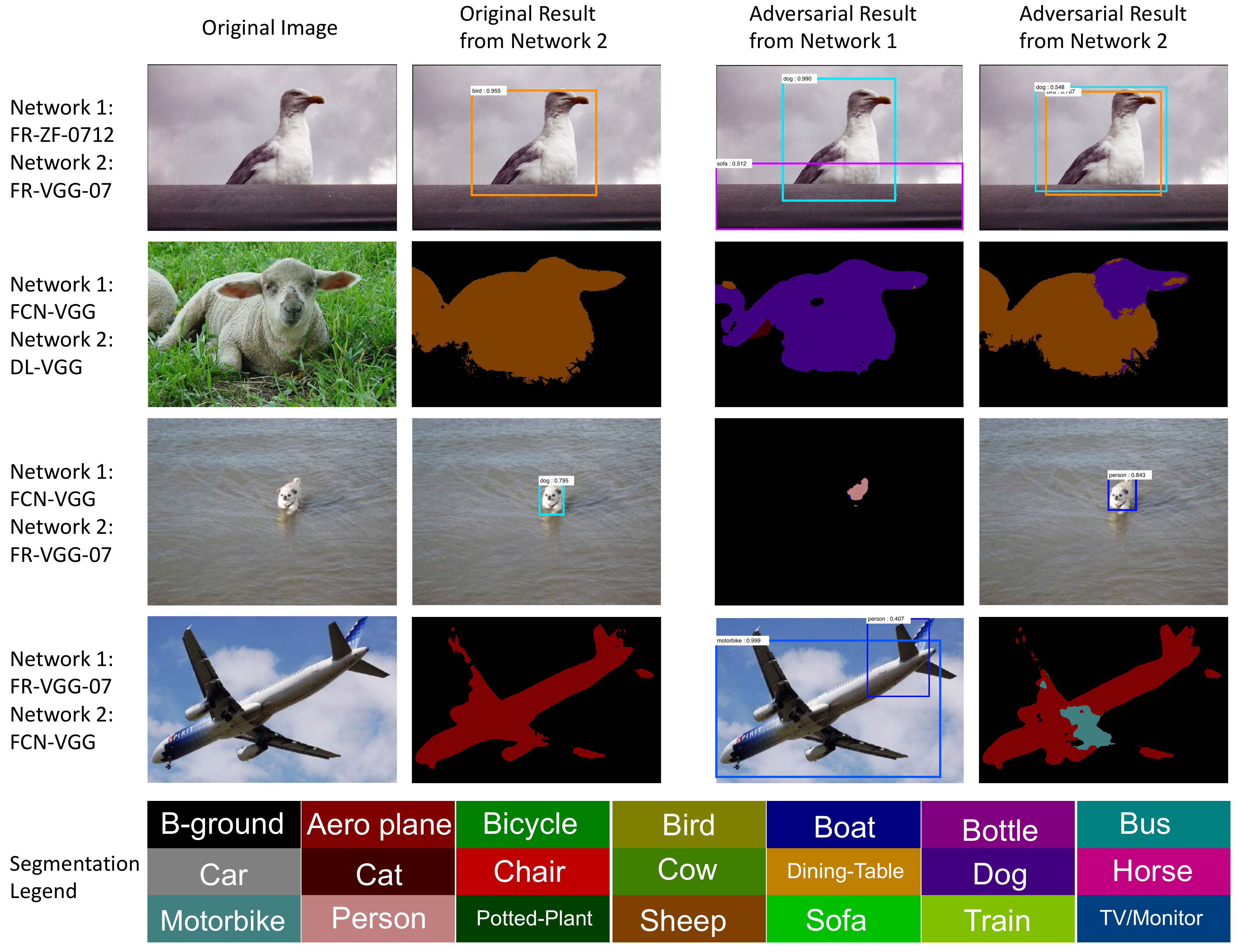}
\vspace{-0.3cm}
\caption{
    Transferrable examples for semantic segmentation and object detection.
    These four rows, from top to bottom, shows the adversarial attack examples
    within two detection networks, within two segmentation networks,
    using a segmentation network to attack a detection network and in the opposite direction, respectively. 
    The segmentation legend borrows from~\cite{zheng2015conditional}.
}
\label{Fig:TransferrableExamples}
\vspace{-0.3cm}
\end{figure*}

In Tables~\ref{Tab:TransferDet}--~\ref{Tab:TransferDetSeg}, we list some results by adding multiple adversarial perturbations.
Also, in order to verify that the spatial structure of combined adversarial perturbations is the key point
that leads to statistically significant accuracy drop,
we randomly generate three permutations of the combined adversarial perturbations and report the average accuracy.
From the results listed in Tables~\ref{Tab:TransferDet}--~\ref{Tab:TransferDetSeg},
we can observe that adding multiple adversarial perturbations often works better than adding a single source of perturbations.
Indeed, the accuracy drop caused by the combined perturbation approximately equals to the sum of drops by each perturbation.
For example, the adversarial perturbation $\mathbf{r}_2+\mathbf{r}_4$ (combining {\bf FR-ZF-0712} and {\bf FR-VGG-0712})
causes significant mAP drop on all {\bf ZFNet}-based and {\bf VGGNet}-based detection networks,
and the adversarial perturbation $\mathbf{r}_5+\mathbf{r}_7$ (combining {\bf FCN-Alex*} and {\bf FCN-VGG*})
causes significant mIOU drop on all {\bf AlexNet}-based and {\bf VGGNet}-based segmentation networks.
However, permutation destroys the spatial structure of the adversarial perturbations, leading to negligible accuracy drops.
The same conclusion holds when the perturbations from different tasks are combined.
Table~\ref{Tab:TransferDetSeg} shows some quantitative results of such combination and Figure~\ref{Fig:FoolAll} shows an example. Note that, the perceptibility value defined in Section~\ref{Generating:Diagnosis:Perceptibility} remains very small even when multiple adversarial perturbations are combine (e.g., $4.0\times10^{-3}$ by $\mathbf{r}_1+\mathbf{r}_3+\mathbf{r}_5+\mathbf{r}_7$).

\subsection{Black-Box Attack}
\label{Transferring:BlackBox}

Combining heterogeneous perturbations allows us to perform better on the so-called {\em black-box attack}~\cite{papernot2016practical},
in which we do not need to know the detailed properties (architecture, purpose, {\em etc.}) about the defender network.
According to the above experiments,
a simple and effective way is to compute the sum of perturbations from several of known networks,
such as {\bf FR-ZF-07}, {\bf FR-VGG-07} and {\bf FCN-Alex}, and use it to attack an unknown network.
This strategy even works well when the structure of the defender is not investigated before.
As an example shown in Table~\ref{Tab:TransferDetSeg}, the perturbation $\mathbf{r}_1+\mathbf{r}_3+\mathbf{r}_5+\mathbf{r}_7$
leads to significant accuracy drop (from $80.20\%$ to $64.52\%$) on {\bf R-FCN-RN101}\cite{li2016r},
a powerful network based on the deep {\bf ResNet}~\cite{he2016deep}.

\section{Conclusions}
\label{Conclusions}

In this paper, we investigate the problem of generating adversarial examples,
and extend it from image classification to semantic segmentation and object detection.
We propose DAG algorithm for this purpose.
The basic idea is to define a dense set of targets as well as a different set of desired labels,
and optimize a loss function in order to produce incorrect recognition results on all the targets simultaneously.
Extensive experimental results verify that DAG is able to generate visually imperceptible perturbation,
so that we can confuse the originally high-confidence recognition results in a well controllable manner.

An intriguing property of the perturbation generated by DAG lies in the transferability.
The perturbation can be transferred across different training sets, different network architectures and even different tasks.
Combining heterogeneous perturbations often leads to more effective adversarial perturbations in black-box attacks.
The transferability also suggests that deep networks, though started with different initialization and trained in different ways,
share some basic principles such as local linearity, which make them sensitive to a similar source of perturbations.
This reveals an interesting topic for future research.

\section*{Acknowledgements}
\label{Acknowledgements}

We thank Dr. Vittal Premachandran, Chenxu Luo, Weichao Qiu, Chenxi Liu, Zhuotun Zhu and Siyuan Qiao for instructive discussions.

\clearpage

{\small
\bibliographystyle{ieee}
\bibliography{egbib}
}

\clearpage

\onecolumn
\appendix

\section{More Fancy Examples}
\label{FancyExamples}

\subsection{Generating Geometric Patterns}
\label{FancyExamples:Geometry}

\begin{figure}[!htb]
\centering
\includegraphics[width=\columnwidth]{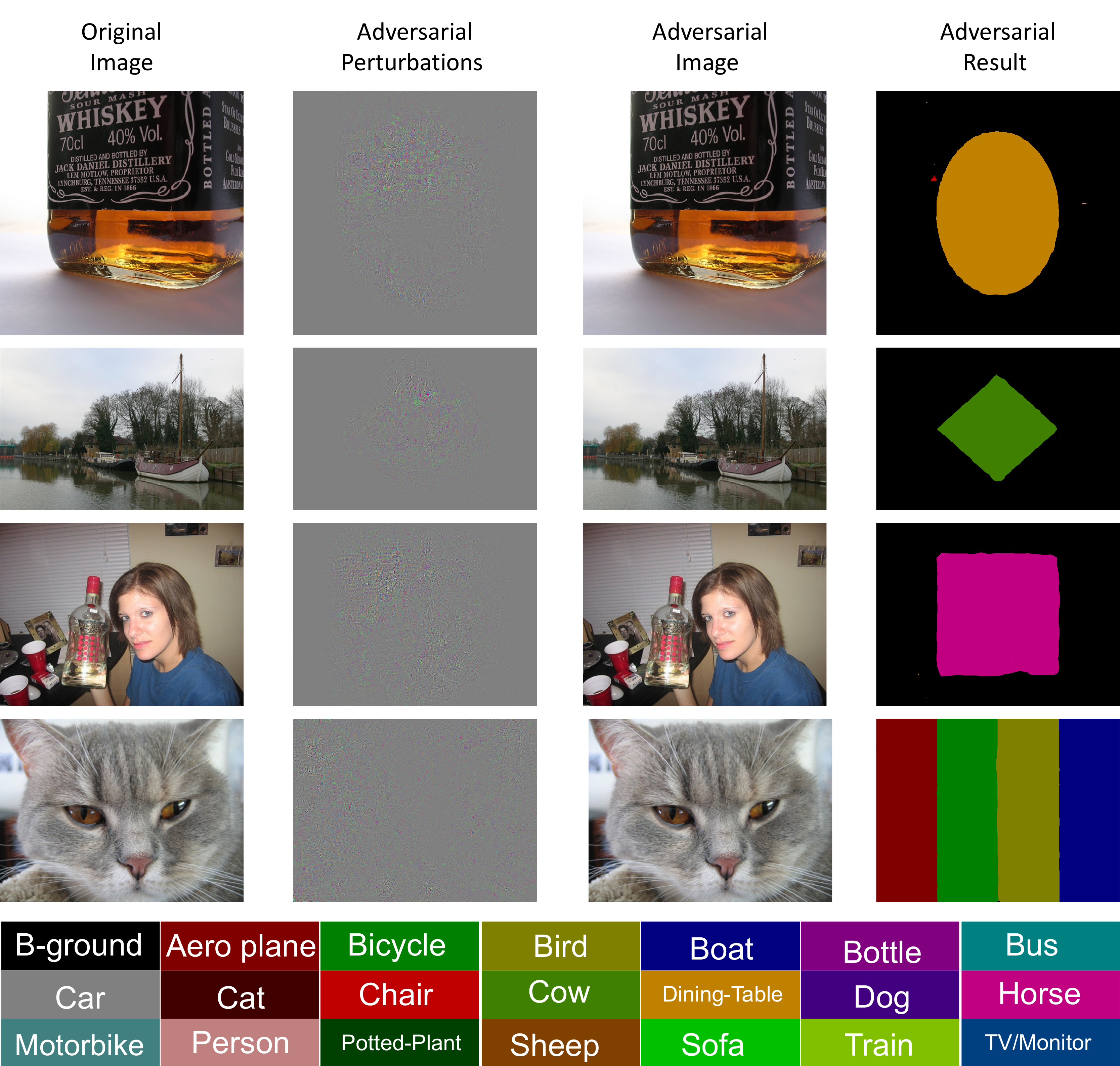}
\caption{
    The adversarial perturbations confuse the deep networks to output different geometric patterns as segmentation results,
    such as a circle (the first row), a diamond (the second row), a square (the third row), and stripes (the fourth row).
    Here, {\bf FCN-Alex} is used as the baseline network (defender).
    All the perturbations are {\bf magnified by $\mathbf{10}$} for better visualization.
}
\vspace{-0.5cm}
\label{Fig:supple_Fig2}
\end{figure}

As an additional showcase, the deep segmentation networks can be confused to output some geometric shapes,
including {\em stripes}, {\em circles}, {\em triangles}, {\em squares}, {\em etc.},
after different adversarial perturbations is added to the original image.
Results are shown in Figure~\ref{Fig:supple_Fig2}.
Here, the added adversarial perturbation varies from case to case.

\subsection{Same Noise, Different Outputs}
\label{FancyExamples:DifferentOutputs}

In Figure~2 of the main article,
we show that we can generate some adversarial perturbations
to make a deep segmentation network output a pre-specified segmentation mask ({\em e.g.}, {\tt ICCV} and {\tt 2017}).
But, the perturbations used to generate these two segmentation masks are different.

\begin{figure}[!htb]
\centering
\includegraphics[width=\columnwidth]{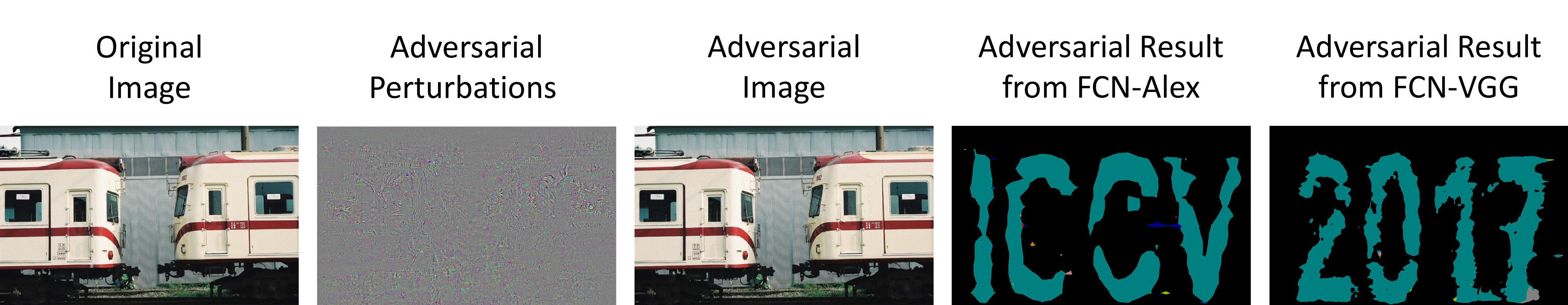}
\caption{
    We add one adversarial perturbation ({\bf magnified by $\mathbf{10}$}) to the same original image
    to generate different pre-specified segmentation masks on two deep segmentation networks ({\bf FCN-Alex} and {\bf FCN-VGG}).
    This is a more difficult task compared to that shown in Figure~2 of the main article,
    where two different adversarial perturbations are used to generate two pre-specified segmentation masks.
    The blue regions in the segmentation masks are predicted as {\em bus}, a randomly selected class.
}
\vspace{-0.5cm}
\label{Fig:supple_Fig1}
\end{figure}

Here, we present a more challenging task, which uses the same perturbations to confuse two networks.
More specifically, we hope to generate a perturbation $\mathbf{r}$, when it is added to an image $\mathbf{X}$,
the {\bf FCN-Alex} and the {\bf FCN-VGG} models are confused to output {\tt ICCV} and {\tt 2017}, respectively.
To implement this, we apply the locally linear property of the network, and add two sources of perturbations,
{\em i.e.}, ${\mathbf{r}}={\mathbf{r}_1+\mathbf{r}_2}$,
where $\mathbf{r}_1$ is generated on {\bf FCN-Alex} with the mask {\tt ICCV},
and $\mathbf{r}_2$ is generated on {\bf FCN-VGG} with the mask {\tt 2017}.
As shown in Figure~\ref{Fig:supple_Fig1}, our simple strategy works very well,
although the segmentation boundary of each letter or digit becomes somewhat jagged.

\end{document}